\newif\ifarxiv
\arxivtrue
 
\ifarxiv
    \documentclass[11pt]{article}
    \usepackage[margin=1in]{geometry}
    \usepackage{times}
    \usepackage{natbib}
    \usepackage{fancyhdr}
 
    \renewcommand{\headrulewidth}{0.4pt}
    \makeatletter
    \renewcommand{\maketitle}{%
      \thispagestyle{fancy}%
      \begin{center}
        \vspace*{0.3em}
        {\LARGE\bfseries \@title\par}
        \vspace{1.2em}
        \@author\par
      \end{center}
      \vspace{0.8em}
    }
    \makeatother
 
    \renewenvironment{abstract}{%
      \begin{center}{\large\scshape Abstract}\end{center}%
      \vspace{-0.3em}%
      \begin{list}{}{\leftmargin=0.4in \rightmargin=0.4in
                     \listparindent=0pt \topsep=0pt \parsep=0.4em}\item\relax
    }{\end{list}\vspace{0.5em}}
\else
    \documentclass{article}
    \usepackage{iclr2026_conference,times}
\fi
 
\usepackage{amsmath,amsfonts,bm}

\def\eqref#1{equation~\ref{#1}}

\def\1{\bm{1}}

\DeclareMathAlphabet{\mathsfit}{\encodingdefault}{\sfdefault}{m}{sl}
\SetMathAlphabet{\mathsfit}{bold}{\encodingdefault}{\sfdefault}{bx}{n}

\usepackage{hyperref}
\usepackage{url}
\usepackage{booktabs}
\usepackage{amsmath,amssymb}
\usepackage{graphicx}
\usepackage{xcolor}
\usepackage{multirow}
\usepackage{algorithm}
\usepackage{amsthm}

\usepackage{algpseudocode}
\algnewcommand\algorithmicinput{\textbf{Input:}}
\algnewcommand\Input{\item[\algorithmicinput]}
\algnewcommand\algorithmicoutput{\textbf{Output:}}
\algnewcommand\Output{\item[\algorithmicoutput]}

\usepackage{listings}
\usepackage{enumitem}
\usepackage{tcolorbox}
\tcbuselibrary{breakable}
   
\usepackage{listings,xcolor}
\definecolor{codebg}{RGB}{247,249,252}
\definecolor{codeframe}{RGB}{200,210,228}
\definecolor{codekw}{RGB}{31,58,147}
\definecolor{codecomment}{RGB}{107,114,128}
\definecolor{codestr}{RGB}{30,132,73}
\lstdefinestyle{scroll}{
  language=Python,
  basicstyle=\ttfamily\small,
  keywordstyle=\color{codekw}\bfseries,
  commentstyle=\color{codecomment}\itshape,
  stringstyle=\color{codestr},
  backgroundcolor=\color{codebg},
  frame=single, rulecolor=\color{codeframe}, framerule=0.4pt,
  xleftmargin=6pt, framexleftmargin=6pt,
  aboveskip=6pt, belowskip=6pt,
  columns=fullflexible, keepspaces=true,
  showstringspaces=false,
}
\lstdefinestyle{traj}{
  style=scroll,
  language={},
  basicstyle=\ttfamily\footnotesize,
  breaklines=true,
  morecomment=[l]{//},
  escapeinside={(*@}{@*)},
}

\definecolor{trajgood}{RGB}{22,101,52}
\definecolor{trajbad}{RGB}{165,42,42}
\newcommand{\trajgoodline}[1]{\textbf{\textcolor{trajgood}{#1}}}
\newcommand{\trajbadline}[1]{\textbf{\textcolor{trajbad}{#1}}}

\hypersetup{
    colorlinks=true,
    linkcolor=black,         
    citecolor=blue!60!black,  
    urlcolor=blue!60!black    
}
 
\newif\ifphfinal
\phfinalfalse  

\ifphfinal
  \newcommand{\ph}[1]{\textbf{??}}
\else
  \newcommand{\ph}[1]{\textcolor{red}{#1}}
\fi

\title{
    Context as an Environment:\\
    Programmatic Context Management for Long-Horizon Agents
}
 
\ifarxiv
    \author{%
      \begin{tabular}[t]{c}
        \textbf{Yin Lin}$^{\dagger}$\\
        {\footnotesize\texttt{yin.lin@alibaba-inc.com}}\\
        {\small Alibaba Group}
      \end{tabular}\hspace{1.2em}
      \begin{tabular}[t]{c}
        \textbf{Elaine Ang}$^{\dagger\,\S}$\\
        {\footnotesize\texttt{ra3448@columbia.edu}}\\
        {\small Columbia University}
      \end{tabular}\hspace{1.2em}
      \begin{tabular}[t]{c}
        \textbf{Erkang Zhu}\\
        {\footnotesize\texttt{erkang.zhu@alibaba-inc.com}}\\
        {\small Alibaba Group}
      \end{tabular}\\[1.2em]
      \begin{tabular}[t]{c}
        \textbf{Bolin Ding}\\
        {\footnotesize\texttt{bolin.ding@alibaba-inc.com}}\\
        {\small Alibaba Group}
      \end{tabular}\hspace{3em}
      \begin{tabular}[t]{c}
        \textbf{Jingren Zhou}\\
        {\footnotesize\texttt{jingren.zhou@alibaba-inc.com}}\\
        {\small Alibaba Group}
      \end{tabular}%
    }
    \date{}
\else
    \author{Anonymous Authors \\
    Anonymous Institution \\
    \texttt{anon@example.com}}
\fi
 
\begin{document}
 
\maketitle
 
\ifarxiv
    \renewcommand{\thefootnote}{}
    \footnotetext{$^{\dagger}$Equal contribution.}
    \footnotetext{$^{\S}$Work done during an internship at Alibaba Group.}
    \renewcommand{\thefootnote}{\arabic{footnote}}
    \setcounter{footnote}{0}
\fi
 
\begin{abstract}
LLM agents increasingly take on long-running tasks whose history grows far beyond a single model context window. Existing approaches compress earlier interactions or extract selected information into fixed memory representations, committing to what to preserve before future needs are known.
We present \emph{Scroll}, a context manager that treats each agent session as an executable \emph{Session Environment}. 
The environment is backed by an append-only Event Log and a sandboxed, persistent Python kernel. The kernel maintains a typed namespace across model calls, allowing tool outputs, retrieved history, and derived state to be bound to variables rather than serialized into the prompt at each call. Model-written code searches, materializes, and transforms session state through \texttt{exec}; only explicitly printed projections enter the model's working view for the next call.
Context management thus becomes a programming task that inherits the improving coding abilities of LLMs, while the Event Log preserves lossless historical ground truth.
As the working view approaches its budget, stale spans are evicted but remain recoverable: an eviction index keeps compact landmarks tied to exact Event Log addresses, so that the agent navigates directly to evicted regions instead of searching the full log.
With Qwen3.8-Max as the backbone, Scroll achieves \textbf{94.8\%} on LongMemEval$_S$; \textbf{73.1\%} on BEAM$_{10M}$, surpassing the best published memory system by 5.1 points; and \textbf{86.7\%} on LOCA$_{256K}$, exceeding the best published long-horizon agent by 37.4 points.
\end{abstract}

\section{Introduction}
\label{sec:intro}

LLM agents are increasingly used for long-running tasks such as repository-level software engineering \citep{jimenez2024swebench,yang2024sweagent} and deep research over the open web \citep{zheng2025deepresearcher,wei2025browsecomp}.
Unlike single-turn generation, these tasks unfold over extended trajectories of model calls, tool executions, observations, failures, and revisions. As trajectories grow, a central challenge for the \emph{agent harness} is context management: session history accumulates continuously, while each model invocation operates over a bounded context window. Moreover, the effective context a model can reliably exploit is far smaller than its nominal window, as long-input retrieval and reasoning degrade with input length \citep{modarressi2025nolima,zeng2026loca}.

Current systems largely address this problem through \emph{context compression} or \emph{external memory}.
Compression is the dominant approach in practice: existing methods truncate stale spans, discard tool outputs, or replace earlier trajectory segments with summaries \citep{kang2025acon,ye2025agentfold}, and production agents such as Claude Code, Codex CLI, and Cursor reportedly employ similar compaction mechanisms as the context window approaches its limit.
External memory systems extract selected facts or episodes into a separate store and later retrieve them through semantic or structured interfaces \citep{packer2023memgpt,wu2025longmemeval, cao2026remember}. 
Both are fundamentally lossy in the same way: the agent sees history only through the summary or the memory store, so any detail they fail to keep is out of reach---even if the raw log still exists on disk.
Long-horizon tasks, however, may require exact historical evidence or nontrivial computation over past events, such as comparing tool outputs produced far apart in the trajectory. Neither the relevant information nor the required operation is known in advance, so no summary produced at observation time can be guaranteed to preserve what is later needed.

We present \emph{Scroll}, which keeps the agent's history outside the model context \citep{zhang2025rlm} and represents it as an executable \emph{Session Environment}. 
An append-only Event Log preserves the interaction trajectory with stable addresses and provenance, while a sandboxed, persistent Python kernel survives across model calls and maintains a typed namespace of resident variables. Any of this state can therefore be materialized as Python objects and reused across reasoning steps without being serialized into the prompt.

This turns context management into \emph{writing programs}---something current models are already highly proficient at. The model issues \texttt{exec} actions to search and expand the Event Log, access permitted resources, invoke tools \citep{ptc}, and compute over resident variables. Retrieved records, tool outputs, and intermediate computations remain in the kernel unless explicitly emitted through \texttt{print}, which the harness inserts as an observation into the next model context. Thus, \texttt{exec} determines how the environment is accessed and transformed, while \texttt{print} determines which projection enters the model's \emph{working view} over the Session Environment.

 As the working view approaches its budget, the harness evicts stale spans. Unlike compaction, eviction changes only the view, never the underlying record: evicted events remain verbatim in the Event Log under their stable addresses, where the model's programs can search for and materialize them on demand. 
Scroll additionally keeps an \emph{eviction index} in the view: a compact map of what has left it. Where search recovers only what the agent thinks to ask for, the index keeps the agent aware of history it can no longer see; each entry anchors the exact addresses of the evicted events, from which the originals are materialized on demand.

Our contributions are threefold:
\begin{itemize}[leftmargin=1.4em,itemsep=1pt,topsep=2pt]
    \item We formulate long-horizon context management as choosing, at each step, a working view over a persistent Session Environment. Existing approaches fix this choice before future needs are known; Scroll defers it to query time as a program the model writes.
    \item We implement an executable context substrate combining an append-only Event Log, durable storage, and a sandboxed persistent Python kernel. The model operates on the environment through \texttt{exec}; within it, only explicit \texttt{print} output crosses into the model-visible context.
    \item We introduce an algorithm that keeps the working view within budget without losing history: evicted spans remain intact in the Event Log, indexed by compact address-anchored entries the agent can navigate directly.
\end{itemize}
\section{Scroll Context Manager}
\label{sec:method}

\begin{figure*}[t]
    \centering
    \includegraphics[width=0.98\textwidth]{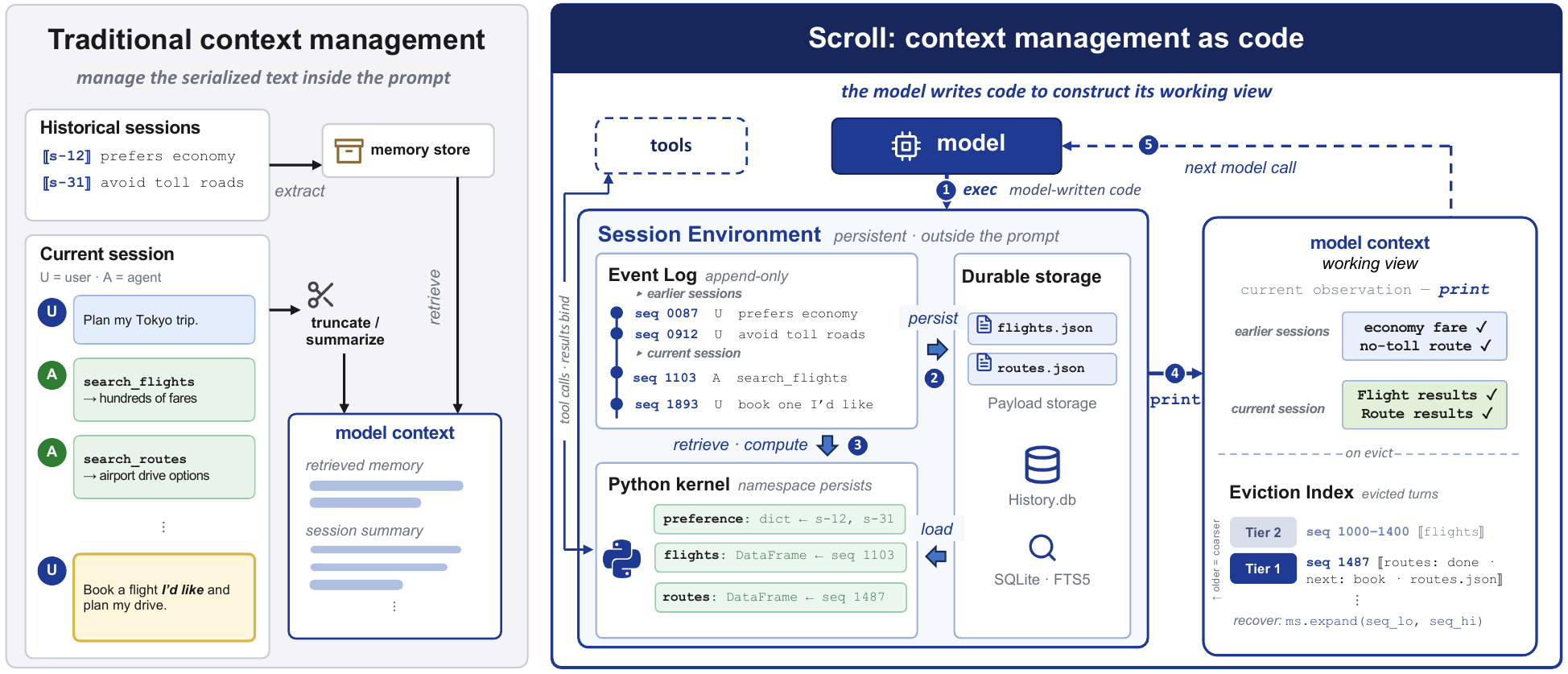}
    \caption{
    Overview of Scroll. Scroll keeps the full session in a persistent, executable Session Environment; model-written code retrieves and computes over it via \texttt{exec}, and \texttt{print} selects the working view exposed to the next model call.
    }
    \label{fig:scroll}
\end{figure*}

We introduce Scroll, a context manager for long-horizon LLM agents. The key insight is that an agent's accumulated history should not be serialized into the model's prompt but should instead be treated as \emph{an environment that the model programmatically interacts with}. The prompt then carries only a working view, while the session lives outside the context window without loss. We first formalize the problem this design addresses (\S\ref{sec:formulation}), then describe the Session Environment (\S\ref{sec:session-environment}), the programmatic interface through which the model constructs its own context (\S\ref{sec:programmatic-context}), and the eviction mechanism that keeps the working view bounded while preserving recoverability (\S\ref{sec:recoverable-eviction}). 

\subsection{Problem Formulation}
\label{sec:formulation}

\paragraph{Session state and working view.}
An agent session produces a growing sequence of \emph{events}
$e_1, e_2, \ldots$ (user messages, model responses, tool calls, tool
results), each with an associated \emph{payload} (its raw content, e.g., a
full tool output). We write the session state after $t$ agent steps as
\begin{equation}
    S_t = (L_t,\; P_t,\; V_t),
\end{equation}
where $L_t$ is the event sequence with per-event metadata, $P_t$ the payloads referenced by $L_t$, and $V_t$ auxiliary derived state (in Scroll, a variable namespace; in other systems, a memory store or summary buffer). 
Each model call, however, consumes a \emph{working view} $c_t$ with $|c_t| \le C$ tokens, where $C$ is the model's nominal context window.
The \emph{context-management problem} is to choose, at every step, the next view: the map $S_t \mapsto c_{t+1}$.

\paragraph{When the selection is made.}
Existing approaches fix the map $S_t \mapsto c_{t+1}$ before future needs are known: compression applies a lossy operator $\phi$ as the trajectory grows, $c_{t+1} = \phi(c_t,\, e_{t})$, deciding which information survives when each segment is compacted; external memory applies an extraction operator $\psi$ at ingestion, $V_{t} = \psi(V_{t-1}, e_{t})$, fixing what is stored and how it can later be retrieved. Either way, the reduced representation replaces the history it summarizes, so anything it omits is unrecoverable.

Scroll instead defers selection to query time. The full state $S_t$ persists losslessly outside the context, and the map $S_t \mapsto c_{t+1}$ is a \emph{program} $\pi_t$ the model writes at step $t$: the program executes on $S_t$, updates $V_t$, and emits a bounded observation for the next call. The model decides what to recall, compute, and expose; the harness makes those decisions safe via durable storage, stable addressing, and sandboxed execution.  Because the policy is expressed as code, it inherits the full generality of programs and improves with the backbone's coding ability, at no change to the harness.

\begin{table}[t]
\centering
\footnotesize
\begin{tabular}{@{}p{0.16\linewidth}p{0.32\linewidth}p{0.43\linewidth}@{}}
\toprule
\textbf{Operation} & \textbf{Interface} & \textbf{Role} \\
\midrule
\textsc{Locate}
&
\texttt{ms.search(query, k, ...)}
&
Find candidate Event Log records via BM25-ranked full-text search and
structured scope, kind, and time filters; each hit carries its stable
\texttt{seq} address.\\
\textsc{Materialize}
&
\texttt{ms.expand(seq)} \newline
\texttt{ms.expand(seq\_lo, seq\_hi)}
&
Recover exact turns or sequence spans, and load externalized payload
contents behind lazy handles. \\
\textsc{Compute}
&
Ordinary Python and permitted database, filesystem, and tool interfaces
&
Filter, join, aggregate, resolve updates, inspect artifacts, or construct
derived state. \\
\textsc{Expose}
&
\texttt{print(value)}
&
Return the selected projection as a bounded observation; everything else
stays in the kernel. \\
\bottomrule
\end{tabular}
\caption{The model-facing interface factorizes context construction into
location, materialization, computation, and exposure.}
\label{tab:memory-surface}
\end{table}

\subsection{Persistent Session Environment}
\label{sec:session-environment}

Scroll realizes $S_t$ as a \emph{Session Environment} (Figure~\ref{fig:scroll}, right), with three components corresponding to $L_t$, $P_t$, and $V_t$.

\paragraph{Append-only Event Log ($L_t$).}
The Event Log is the durable, ground-truth record of an agent's sessions: a single append-only log that spans session boundaries. Every interaction appends a typed event carrying the metadata future queries need---role, session and agent identifiers, timestamps, and tool state---and receives its immutable, monotonically increasing \texttt{seq}. Our implementation stores events in SQLite. Search defaults to BM25 rather than embeddings: it is deterministic and requires no index-time model calls. 

\paragraph{Durable storage ($P_t$).}
A payload is the raw content an interaction produced, such as a full tool result or a generated artifact. The log records that the interaction occurred but need not store every byte of it in the event row: small payloads remain inline in SQLite, while large ones are moved into JSON or artifact storage on the filesystem, with the row retaining a bounded preview and a recovery pointer. Externalized payloads are accessed through lazy handles (\texttt{ToolResultRef}, \texttt{ArtifactRef}).

\paragraph{Persistent runtime and resident namespace ($V_t$).}
A sandboxed Python kernel persists across model calls throughout the session; its namespace holds \emph{environment objects}: resident Python values and lazy handles, each carrying type, size, and provenance metadata identifying the events it derives from. 
Tool invocations issued through the programmatic tool interface~\citep{ptc} return Python objects that later programs can operate on.
The harness prepends to every call a \emph{namespace digest}: a short listing of each resident variable's name, type, and shape, with small scalar values shown inline. Model-authored code runs in a fail-closed sandbox: the Event Log is read-only from the kernel, and database, filesystem, network, and tool access are limited to capabilities the harness explicitly declares.

\begin{figure*}[t]
    \centering
    \includegraphics[width=0.7\textwidth]{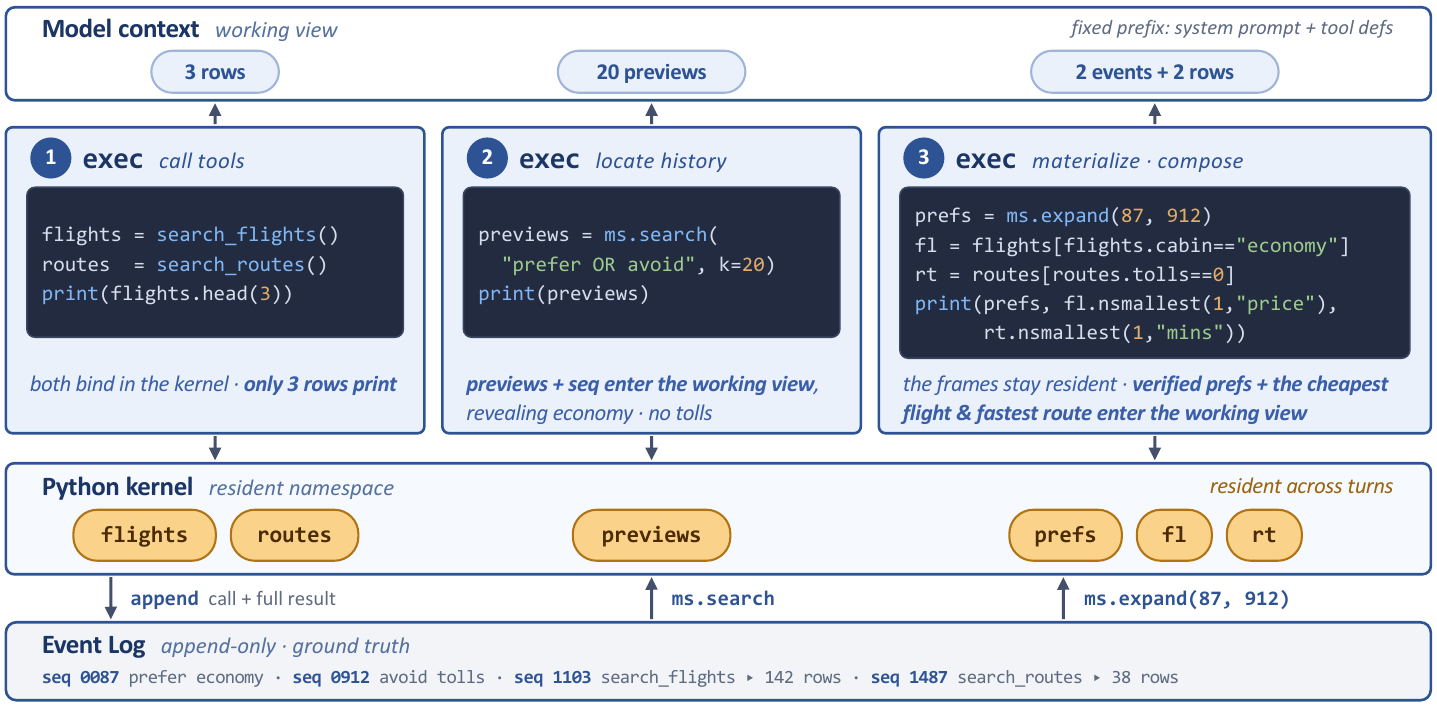}
    \caption{
    Programmatic context construction. Three \texttt{exec} turns compute over resident state in the kernel; only \texttt{print} output crosses into the next model context/working view. 
    }
    \label{fig:dataflow}
\end{figure*}

\subsection{Programmatic Context Construction}
\label{sec:programmatic-context}

Scroll uses a CodeAct-style interface~\citep{codeact} for both task execution and context construction. A controlled capability object, \texttt{ms}, forms the model-facing \emph{memory surface} over durable history, abstracting the physical backend behind four operations (Table~\ref{tab:memory-surface}).

Figure~\ref{fig:dataflow} traces the trip-planning task of Figure~\ref{fig:scroll} through three \texttt{exec} cells. Cell~1 binds full tool results to the resident variables \texttt{flights} and \texttt{routes} in the Python kernel, printing only a few rows. Cell~2 searches the Event Log for stated preferences; the matching previews and \texttt{seq} addresses enter the working view, revealing the user's preference for economy cabins and toll-free routes.
Cell~3 expands the two events for the verbatim record, filters and ranks the resident variables accordingly, and prints the two preference turns alongside the cheapest economy flight and the fastest toll-free route. The bulk tool results never enter the working view; every call is appended to the Event Log with its full result, addressable by \texttt{seq}.

\subsection{Eviction and Off-Context Navigation}
\label{sec:recoverable-eviction}

The working view must stay bounded as the session grows. Scroll bounds it with an eviction procedure (Algorithm~\ref{alg:eviction}) that triggers whenever the working view exceeds a budget $\rho C$. 
The procedure first persists any live turns to the Event Log and protects the active turn, the recent tail, and the newest tool results. The remainder is evicted in increasing order of recovery cost: completed tool payloads are folded first, since a single \texttt{seq} pointer suffices to recover them; whole spans are removed only if the view remains over budget.
What leaves the view is not lost: it stays verbatim in the Event Log, and the procedure's one invariant is that everything it removes stays addressable.

\begin{algorithm}[t]
\caption{Recoverable context eviction}
\label{alg:eviction}
\begin{algorithmic}[1]
\Input working view $c$, Event Log $\mathcal{L}$, eviction index $\mathcal{I}$, budget $\rho C$, tier width $k$
\Output bounded $c$ and updated $\mathcal{I}$; removed spans stay recoverable from $\mathcal{L}$
\If{$|c| > \rho C$}
  \State $\mathcal{L} \gets \Call{Persist}{c, \mathcal{L}}$
    \Comment{live turns become durable}
  \State $R \gets \Call{Protected}{c}$
    \Comment{active turn, recent tail, newest tool results}
  \State $c \gets R \,\cup\, \Call{FoldPayloads}{c \setminus R}$
    \Comment{payloads $\to$ \texttt{seq} pointers}
  \State $E \gets \Call{SelectSpan}{c \setminus R,\; |c| - \rho C}$
    \Comment{oldest completed span above budget}
  \State $c,\ \mathcal{I} \gets \Call{EvictToIndex}{c,\; E,\; \mathcal{H}[E]}$
    \Comment{$E$ leaves the view; its headlines enter $\mathcal{I}$, shown in place}
  \State $\mathcal{I} \gets \Call{RollUp}{\mathcal{I},\, k}$
\EndIf
\State \Return $(c, \mathcal{I})$
\end{algorithmic}
\end{algorithm}

\paragraph{Headlines as navigation anchors.}
Lexical search recovers an evicted span only when the agent recalls its wording; Scroll therefore also maintains \emph{landmarks} for position-based navigation. As part of each response, the model writes a short \emph{headline}---task, verified state, next action, and a status---which Scroll binds at append time to the \texttt{seq} assigned by the Event Log, yielding a map $\mathcal{H}$ from address to headline. When a span is evicted, its headlines enter a tiered index (Figure~\ref{fig:scroll}). 
A flat index would grow linearly with the session, so Scroll rolls it up: each tier holds at most $k$ blocks; when a tier fills, the newest block retains full detail while the $k-1$ older ones collapse to one line each and merge into the next tier.
After $n$ evictions, the index occupies $O(k \log_k n)$ blocks, providing fine anchors for recent history, coarse ranges for distant history, each backed by a \texttt{seq} span.
\section{Experimental Setup}
\label{sec:setup}

\subsection{Benchmarks}
\label{sec:setup-benchmarks}

We evaluate Scroll in two long-horizon settings:
(1) retrieving and reasoning over interaction histories that exceed the live context, and (2) reasoning and acting in an agentic environment whose state grows over time.

\paragraph{Long-term memory retrieval and reasoning.}
\texttt{LongMemEval}~\citep{wu2025longmemeval} poses questions over a history of prior user--assistant conversations. Answering may require locating evidence scattered across sessions, resolving temporal dependencies and knowledge updates, and reasoning over the retrieved evidence. The benchmark provides three settings with increasing amounts of distractor history: \texttt{Oracle}, where the history contains only the evidence sessions; and \texttt{S} and \texttt{M}, where each question is paired with roughly 50 and 500 sessions (${\sim}115$K and ${\sim}1.5$M tokens) of history, respectively.

\texttt{BEAM}~\citep{beam} extends this evaluation to substantially longer coherent histories. Its questions may require collecting non-adjacent evidence, tracking changes over time, deduplicating repeated information, or aggregating facts distributed throughout the history. The benchmark spans four history scales (128K, 500K, 1M, and 10M tokens); at the largest scale, BEAM$_{10M}$, histories cannot be consumed directly within current model context windows.

\paragraph{Long-context reasoning and acting.}
\texttt{LOCA}~\citep{zeng2026loca} evaluates agents that must reason, invoke tools, and modify an environment as the available environment state accumulates. LOCA measures whether an agent can continue to reason and act throughout a growing tool-use trajectory. The benchmark scales the \emph{environment description length} (the token count of the full environment state as seen through tool outputs) across seven regimes from 8K to 256K tokens; we evaluate on the two largest regimes (128K and 256K).

\subsection{Agent Configuration}
\label{sec:setup-agent}

Our main experiments use Qwen3.8-Max as the agent backbone. All context management methods are implemented and evaluated on top of QwenPaw~\citep{qwenpaw}, an agent operating system providing tool invocation and execution infrastructure, and orchestrated with Harbor~\citep{harbor} in the benchmark-provided environments. Our implementation to reproduce all reported results is available at \url{https://github.com/niceIrene/QwenPaw/tree/scroll-research}.

Scroll exposes its functionality to the agent through a set of tools, of which the following two implement the \texttt{exec} action of Section~\ref{sec:method}.

\begin{itemize}[leftmargin=1.4em,itemsep=1pt,topsep=2pt]
    \item \texttt{repl\_exec} executes a model-generated Python cell in the persistent kernel. Environment tools are exposed as Python functions forwarded to the underlying services, enabling programmatic tool calling; all intermediate computation stays in the kernel, and only explicit, budgeted \texttt{print} output enters the model's working window.
    \item \texttt{recall\_history\_python} executes a cell with the memory surface \texttt{ms} bound: \texttt{ms.search} locates evicted records and \texttt{ms.expand} materializes them as Python objects, which the model filters, combines, or aggregates in the kernel before printing a distilled result.
\end{itemize}

We use a single system prompt and one set of context-management rules across all benchmarks, with no few-shot demonstrations. Each memory benchmark contributes only a short rubric specifying its data layout, memory-surface usage, evidence-selection conventions, and answer format (see detailed prompts in Appendix~\ref{app:prompts}). For LOCA, we use the benchmark's official task instructions unmodified, adding only environment metadata (available APIs and workspace paths). 

To test generality across backbones, we additionally evaluate Qwen3.7-Max, Deepseek-v4-pro, GLM-5.2, Kimi-K2.7, and Qwen3.6-35B-A3B (an open-weight model with a smaller active-parameter footprint), changing only the foundation model.

\subsection{Evaluation Protocol}
\label{sec:setup-protocol}

For the two memory benchmarks, we ingest each conversation history into Scroll session by session, in chronological order. At each session boundary, the raw context is cleared, and only Scroll's internal state (the eviction index and the Event Log) is carried forward to subsequent sessions.
For LOCA, each task starts from the benchmark-provided initial environment state. The agent explores and acts on the environment directly, with Scroll managing its context as the trajectory grows.

LongMemEval and BEAM are scored with their benchmark-provided LLM-as-a-judge prompts, using Qwen3.6-flash at temperature \(0\) as the judge; we report accuracy for LongMemEval and the judge score for BEAM. LOCA is scored with its native rule-based verifier, which checks the final environment state, and we report accuracy. Unless otherwise noted, each task is evaluated once in the benchmark-provided container with a random seed; we also record model-facing input and output tokens and the number of interaction turns for each task.

\section{Results}
\label{sec:results}

\subsection{Comparison with Existing Systems}
\label{sec:results-main}

\paragraph{Retrieval accuracy comparison with long-term memory systems
(Table~\ref{tab:memory-results}).}
Agents do not natively retain information across sessions, so answering questions over prior interactions requires an external memory system. 
We compare Scroll against dedicated long-term memory systems. For each system, we report the best publicly available result under its own preferred configuration (backbone model, retrieval budget, and judge) as of August~15, 2026.\footnote{We do not reproduce the baselines ourselves, as independent reproductions in this area have repeatedly led to disagreement over evaluation setup~\citep{zep-locomo-blog, mem0-zep-issue}.} 
Appendix~\ref{app:breakdown} provides per-category breakdowns of Scroll on the \texttt{S} and \texttt{M} splits of LongMemEval and on BEAM$_{10M}$.

\begin{table}[t]
\centering
\caption{Comparison with existing long-term memory systems. For each baseline, we report the best publicly available result under that system's own setup as of August~15, 2026; ``–'' denotes no publicly reported result. These are reference points from the literature rather than a controlled comparison: reader models differ across rows and can substantially affect scores---EmergenceMem (GPT-4o), Zep (GPT-5.4), Mastra OM (GPT-5 mini), Mem0 (GPT-5), Hindsight (Gemini~3~Pro), Exabase M-1 (Gemini~3~Flash), RAG and LIGHT (Llama-4-Maverick); Honcho uses a multi-model pipeline and Cognee does not
report its reader.}
\label{tab:memory-results}
\small
\begin{tabular}{lcc}
\toprule
Method & LongMemEval$_S$ & BEAM$_{10M}$ \\
\midrule
RAG~\citep{beam}                     & --   & 24.9 \\
LIGHT~\citep{beam}                   & --   & 26.6 \\
\midrule
Zep~\citep{zep-research}             & 90.2 & --   \\
Mem0~\citep{mem0,mem0-eval}          & 94.4 & 48.6 \\
Hindsight~\citep{hindsight}          & 94.6 & 64.1 \\
EmergenceMem~\citep{emergencemem}    & 86.0 & --   \\
Honcho~\citep{honcho}                & 90.4 & 40.6 \\
Mastra OM~\citep{mastra}             & 94.9 & --   \\
Cognee~\citep{cognee-beam}           & --   & 67.0 \\
Exabase M-1~\citep{exabase}          & \textbf{96.4} & 68.0 \\
\midrule
Scroll (ours)                        & 94.8 & \textbf{73.1} \\
\bottomrule
\end{tabular}
\end{table}

As shown in Table \ref{tab:memory-results}, Scroll is competitive with the strongest reported systems on LongMemEval$_S$ and beats the best-performing system by 5.1 points on BEAM$_{10M}$. Existing memory systems follow a three-stage paradigm: at ingestion, an LLM processes the history into a derived store through fact extraction, summarization, or knowledge-graph construction; at query time, a retrieval pipeline selects candidate memories from that store; and a reader model then reasons over the returned snippets to produce the answer. Scroll instead ingests the raw history as-is, composes retrieval code per question, and needs no separate reader: the agent that writes and executes the queries also produces the final answer directly. 

\begin{table}[t]
\centering
\caption{LOCA accuracy (\%) of different context-management strategies at
the two largest environment description lengths. All agent loops use
Qwen3.8-Max as the backbone;
$\Delta$ denotes the absolute drop from 128K to 256K.}
\label{tab:loca-results}
\small
\begin{tabular}{lccc}
\toprule
Agent loop & 128K & 256K & $\Delta$ \\
\midrule
Summarization Agent            & 86.7 & 65.3 & -21.4 \\
Retrieval Agent           & 88.0 & 66.7 & -21.3 \\
CodeAct Agent                  & \textbf{89.3} & 85.3 & -4.0 \\
\midrule
Scroll (ours)                  & \textbf{89.3} & \textbf{86.7} & \textbf{-2.6} \\
\bottomrule
\end{tabular}
\end{table}

\paragraph{Comparison of context-management strategies on LOCA (Table~\ref{tab:loca-results}).}
On LOCA, we compare four agents that share the same backbone (Qwen3.8-Max) and toolset, and differ only in how they manage a growing context: (i) a \emph{summarization agent}, a ReAct agent~\citep{react} that periodically compacts its interaction history into a summary; (ii) a \emph{retrieval agent}, a ReAct agent whose overflowing history is evicted and made accessible through a recall tool; (iii) a \emph{CodeAct agent}~\citep{codeact} that interacts with the environment through programmatic tool calling; and (iv) Scroll. A comparison against the best published numbers from the LOCA paper~\citep{zeng2026loca} and its leaderboard is in Appendix~\ref{app:loca-full}.

Table~\ref{tab:loca-results} reports accuracy at the two largest environment description lengths. The CodeAct agent and the agent with Scroll, which bind intermediate results to environment objects instead of carrying raw text in context, achieve the best performance and the smallest decrease as context grows.

\subsection{Can Different Backbone Models Use Scroll Effectively?}
\label{sec:results-backbone}

Scroll provides an environment for managing context but does not dictate its use: what state to keep where, and what code to write, are left to the model. A natural question is \emph{whether the ability to use Scroll effectively is specific to one backbone or shared across models of varying capability}. We rerun both regimes across six backbones with the harness, tools, prompts, and context-management rules held fixed (Table~\ref{tab:backbone}).

Every backbone can use Scroll, but stronger models benefit more. On LongMemEval$_S$, where the model queries the history database with short programs, all backbones benefit similarly---even the 35B model reaches $88.8$, within six points of the best ($94.8$). On BEAM$_{10M}$ the gap stays within $15$ points. 
On LOCA, however, tasks demand longer trajectories and
more complex program synthesis, so the spread widens to $64$ points at 256K
($86.7$ vs.\ $22.7$). Failures are not protocol-level: all backbones adhere to the CodeAct interface, but weaker models commit more execution errors or terminate prematurely on aggregation-heavy tasks. Scroll's ceiling on such tasks thus rises with multi-step query planning and the ability to decide when evidence suffices, suggesting room for post-training on frontier model traces.

\begin{table}[t]
\centering
\caption{Scroll across backbones. Only the foundation model changes; harness,
tools, prompts, and context-management rules are identical.}
\label{tab:backbone}
\small
\begin{tabular}{lcccc}
\toprule
Backbone & LongMemEval$_S$ & BEAM$_{10M}$ & LOCA 128K & LOCA 256K \\
\midrule
Qwen3.8-Max      & 94.8 & 73.1 &89.3 & 86.7 \\
Qwen3.7-Max      & 92.8 & 66.6 &78.7 & 60.0 \\
Deepseek-v4-pro  & 93.2 & 70.2 &69.3 & 58.7 \\
GLM-5.2          & 93.6 & 70.7 &66.7 & 62.7 \\
Kimi-K2.7        & 92.0 & 67.2 &30.7 & 32.0 \\
Qwen3.6-35B-A3B  & 88.8 & 58.1 & 37.3 & 22.7 \\
\bottomrule
\end{tabular}
\end{table}


\subsection{Ablation Study}
\label{sec:ablation}
\begin{figure}[t]
\centering
\includegraphics[width=0.8\textwidth]{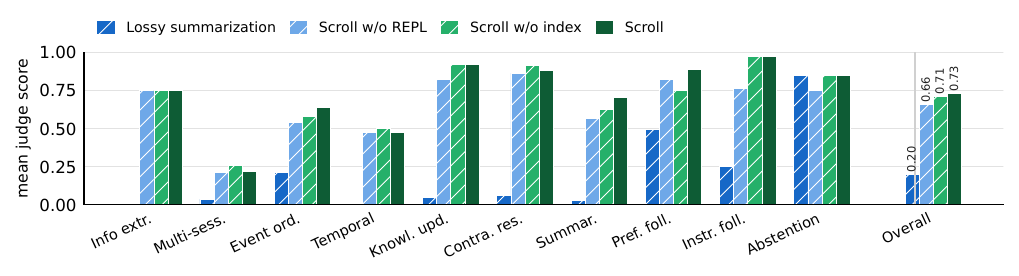}
\vspace{-2mm}
\caption{Ablating Scroll's components on BEAM$_{\text{10M}}$ (judge scores; hatched bars are Scroll variants; Qwen3.8-Max, thinking on). Lossy summarization: summaries replace the originals at ingestion. Scroll w/o REPL: \texttt{ms} exposed as ordinary tool calls, with no persistent kernel. Scroll w/o index: the eviction index is removed, leaving keyword search only. Scroll: the full system.}
\label{fig:ablation}
\end{figure}

We ablate the core components of Scroll on BEAM$_{10M}$. 
Figure~\ref{fig:ablation} reports judge scores per category and overall.
First, to assess the utility of the Event Log, we compare Scroll against a lossy variant whose history is summarized at ingestion, with the originals discarded. 
Second, to evaluate the programmatic interface, we compare against \emph{Scroll w/o REPL}, which exposes \texttt{search}, \texttt{expand}, and \texttt{sql\_query} as ordinary tool calls, with no persistent kernel. 
Third, to assess index-guided navigation, we remove the eviction index, leaving the agent to locate history through keyword search alone rather than index ranges.

Discarding the original records is the most damaging ablation: the lossy variant falls to 19.9 overall, with near-zero scores wherever the answer must preserve exact values from the history, such as information extraction, temporal reasoning, and knowledge update. 
Scroll w/o REPL underperforms full Scroll by 7.3 points, since serialized tool results cannot be filtered, joined, or aggregated in the kernel; the difference is concentrated in abilities that require composing evidence from many records, such as knowledge update (92.5 vs.\ 82.5) and instruction following (97.5 vs.\ 76.3), while single-lookup abilities are unaffected. 
Removing the eviction index costs 1.8 points overall, but the effect concentrates where evidence is scattered across the history and must otherwise be collected by keyword search: preference following (89.1 vs.\ 74.9), summarization (70.5 vs.\ 62.6), and event ordering (64.1 vs.\ 58.1).

\subsection{Cost and Efficiency}
\label{sec:results-cost}
\begin{figure*}[t]
\centering
\includegraphics[width=0.9\textwidth]{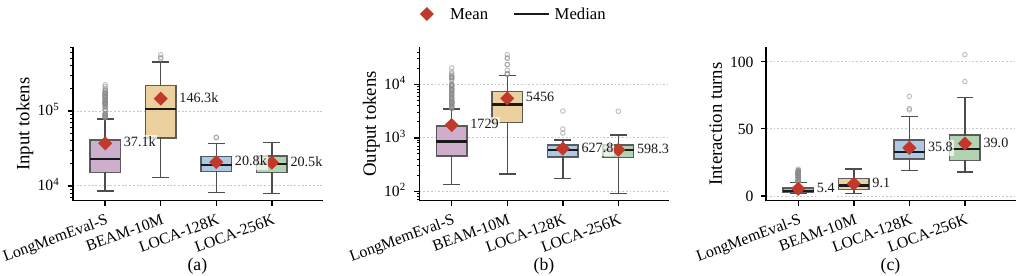}
\caption{Per-task cost of Scroll (backbone: Qwen3.8-Max): (a) input tokens, (b) output tokens, (c) agent turns. Boxes span the IQR with the median marked; red diamonds are means. Log scale in (a) and (b).}
\vspace{-2mm}
\label{fig:cost}
\end{figure*}

Scroll exposes only a small fraction of the corpus to the model. Ingestion involves no additional LLM calls, and at query time records are filtered inside the Python kernel, so only printed output enters the context. Figure~\ref{fig:cost} shows the per-task distribution of input tokens, output tokens, and agent turns: median input on BEAM$_{10M}$ is $105$K tokens, about $1\%$ of the corpus, and output is an order of magnitude smaller than input across all three benchmarks. Note that for LongMemEval$_S$ and BEAM$_{10M}$ we measure retrieval alone, whereas for LOCA we measure full task completion, hence its longer trajectories. We report token counts rather than latency or dollar cost, as both depend on serving configuration.
 
\section{Related Work}
\label{sec:related}

\paragraph{Context compression and external memory.}
Most context-management systems either compress the active trajectory or store selected information externally. Compression methods summarize, clear, or fold earlier interactions into shorter representations \citep{kang2025acon,ye2025agentfold,zhou2026mem1,kontonis2026memento}.
External-memory systems instead extract facts, episodes, or notes into a separate store and retrieve them when relevant \citep{packer2023memgpt,tan2026lightmem,letta2026contextrepos}. 
Both approaches reduce the history the model sees by deciding, before future needs are known, which information survives, in what form, and through which interface it can later be reached.
Scroll instead retains the original interaction events and referenced payloads; summaries and indexes provide compact working views without becoming the sole representation of historical evidence.

\paragraph{Code as the agent–environment interface.}
CodeAct introduced executable Python as a general action interface for LLM agents \citep{codeact}, while programmatic tool-calling and code-execution systems allow tool results to remain in sandbox variables and enter context only through selected projections \citep{ptc,anthropic2025codeexec}. Related work has also explored programmatic access to externalized long input prompts \citep{zhang2025rlm} and structured working state \citep{li2026userascode,vista2026}. Scroll applies this principle to the continuously evolving state of an agent session. Its persistent Python kernel retains typed variables across model calls: \texttt{exec} retrieves and transforms session state, while only explicit \texttt{print} outputs cross the observation boundary.

\paragraph{Lossless session history and navigation.}
Prior systems have explored verbatim recall storage, event-sourced interaction logs, lossless pointers, and provenance-linked memory \citep{packer2023memgpt,nakajima2026log,ehrlich2026lcm,zhang2026mandol}.
Scroll combines a queryable append-only Event Log with external payload references and an executable resident namespace. Its within-session eviction index is a navigation layer over this retained state: recent history is represented by fine-grained, sequence-addressed headlines, while older history is represented by coarser ranges. Once a relevant region is located, the original events and payloads are recovered programmatically. In the terminology of CoALA \citep{sumers2024coala} and context-engineering surveys \citep{mei2025survey}, Scroll connects executable working state with verbatim episodic history through a persistent Session Environment.

\section{Conclusion} \label{sec:conclusion} 

In this report, we present Scroll, a context manager that makes context management an explicit model policy over a persistent Session Environment: the model uses \texttt{exec} to retrieve and compute over externalized state, and uses \texttt{print} to decide what enters the next context, while the harness provides deterministic storage, execution, and recovery. This policy can in turn be distilled from frontier models into smaller ones. Successful trajectories supervise two decisions: \emph{context retrieval} (when and how to write retrieval code over the agent history) and \emph{context injection} (which computed results should be printed back into the working window). We plan to use frontier-model traces for supervised fine-tuning or policy distillation, keeping the underlying context mechanisms fixed.

\ifarxiv
    \bibliographystyle{plainnat}
\else
    \bibliographystyle{iclr2026_conference}
\fi
\bibliography{refs}

@article{zeng2026loca,
  title={Loca-bench: Benchmarking language agents under controllable and extreme context growth},
  author={Zeng, Weihao and Huang, Yuzhen and He, Junxian},
  journal={arXiv preprint arXiv:2602.07962},
  year={2026}
}

@article{kang2025acon,
  title={Acon: Optimizing context compression for long-horizon llm agents},
  author={Kang, Minki and Chen, Wei-Ning and Han, Dongge and Inan, Huseyin A and Wutschitz, Lukas and Chen, Yanzhi and Sim, Robert and Rajmohan, Saravan},
  journal={arXiv preprint arXiv:2510.00615},
  year={2025}
}

@inproceedings{zhou2026mem1,
  title     = {{MEM1}: Learning to Synergize Memory and Reasoning for Efficient Long-Horizon Agents},
  author    = {Zhou, Zijian and Qu, Ao and Wu, Zhaoxuan and Kim, Sunghwan and Prakash, Alok and Rus, Daniela and Low, Bryan Kian Hsiang and Liang, Paul Pu},
  booktitle = {International Conference on Learning Representations (ICLR)},
  year      = {2026}
}

@article{zhang2025rlm,
  title   = {Recursive Language Models},
  author  = {Zhang, Alex L. and Kraska, Tim and Khattab, Omar},
  journal = {arXiv preprint arXiv:2512.24601},
  year    = {2025}
}

@article{ehrlich2026lcm,
  title   = {{LCM}: Lossless Context Management},
  author  = {Ehrlich, Clint and Blackman, Theodore},
  journal = {Voltropy PBC technical report},
  year    = {2026},
  howpublished   = {\url{https://papers.voltropy.com/LCM}}
}

@misc{anthropic2025codeexec,
  title        = {Code Execution with {MCP}: Building More Efficient Agents},
  author       = {{Anthropic}},
  year         = {2025},
  howpublished = {\url{https://www.anthropic.com/engineering/code-execution-with-mcp}}
}

@inproceedings{wu2025longmemeval,
  title     = {{LongMemEval}: Benchmarking Chat Assistants on Long-Term Interactive Memory},
  author    = {Wu, Di and Wang, Hongwei and Yu, Wenhao and Zhang, Yuwei and Chang, Kai-Wei and Yu, Dong},
  booktitle = {International Conference on Learning Representations (ICLR)},
  year      = {2025}
}

@inproceedings{modarressi2025nolima,
  title     = {{NoLiMa}: Long-Context Evaluation Beyond Literal Matching},
  author    = {Modarressi, Ali and Deilamsalehy, Hanieh and Dernoncourt, Franck and Bui, Trung and Rossi, Ryan A. and Yoon, Seunghyun and Sch{\"u}tze, Hinrich},
  booktitle = {International Conference on Machine Learning (ICML)},
  year      = {2025}
}

@article{packer2023memgpt,
  title   = {{MemGPT}: Towards {LLM}s as Operating Systems},
  author  = {Packer, Charles and Fang, Vivian and Patil, Shishir G. and Lin, Kevin and Wooders, Sarah and Gonzalez, Joseph E.},
  journal = {arXiv preprint arXiv:2310.08560},
  year    = {2023}
}

@article{ye2025agentfold,
  title={Agentfold: Long-horizon web agents with proactive context management},
  author={Ye, Rui and Zhang, Zhongwang and Li, Kuan and Yin, Huifeng and Tao, Zhengwei and Zhao, Yida and Su, Liangcai and Zhang, Liwen and Qiao, Zile and Wang, Xinyu and others},
  journal={arXiv preprint arXiv:2510.24699},
  year={2025}
}

@article{zhang2026mandol,
  title   = {{Mandol}: An Agglomerative Agent Memory System for Long-Term Conversations},
  author  = {Zhang, Yuhan and Guo, Zhiyuan and Zeng, Ziheng and Wang, Wei and Wu, Wentao and Xu, Lijie},
  journal = {arXiv preprint arXiv:2606.29778},
  year    = {2026}
}

@inproceedings{tan2026lightmem,
  title     = {A Lightweight, Domain-Adaptive Memory System for {LLM} Agents},
  author    = {Tan, Juntao and Yang, Liangwei and Zhao, Wenting and Qiu, Jielin and Zhu, Ming and Murthy, Rithesh and Savarese, Silvio and Wang, Huan and Heinecke, Shelby and Xiong, Caiming},
  booktitle = {International Conference on Learning Representations (ICLR)},
  year      = {2026},
}

@article{sumers2024coala,
  title   = {Cognitive Architectures for Language Agents},
  author  = {Sumers, Theodore R. and Yao, Shunyu and Narasimhan, Karthik and Griffiths, Thomas L.},
  journal = {Transactions on Machine Learning Research (TMLR)},
  year    = {2024}
}

@inproceedings{yang2024sweagent,
  title     = {{SWE}-agent: Agent-Computer Interfaces Enable Automated Software Engineering},
  author    = {Yang, John and Jimenez, Carlos E. and Wettig, Alexander and Lieret, Kilian and Yao, Shunyu and Narasimhan, Karthik and Press, Ofir},
  booktitle = {Advances in Neural Information Processing Systems (NeurIPS)},
  year      = {2024}
}

@misc{mei2025survey,
      title={A Survey of Context Engineering for Large Language Models}, 
      author={Lingrui Mei and Jiayu Yao and Yuyao Ge and Yiwei Wang and Baolong Bi and Yujun Cai and Jiazhi Liu and Mingyu Li and Zhong-Zhi Li and Duzhen Zhang and Chenlin Zhou and Jiayi Mao and Tianze Xia and Jiafeng Guo and Shenghua Liu},
      year={2025},
      eprint={2507.13334},
      archivePrefix={arXiv},
      primaryClass={cs.CL},
      url={https://arxiv.org/abs/2507.13334}, 
}

@article{li2026userascode,
  title   = {User as Code: Executable Memory for Personalized Agents},
  author  = {Li, Bojie},
  journal = {arXiv preprint arXiv:2606.16707},
  year    = {2026}
}

@article{nakajima2026log,
  title={The Log is the Agent: Event-Sourced Reactive Graphs for Auditable, Forkable Agentic Systems},
  author={Nakajima, Yohei},
  journal={arXiv preprint arXiv:2605.21997},
  year={2026}
}

@article{vista2026,
  title   = {{LLM} Agents Are Latent Context Managers: Typed Working Memory and State Proprioception},
  author  = {{VISTA}},
  journal = {arXiv preprint arXiv:2606.30005},
  year    = {2026}
}

@misc{letta2026contextrepos,
  title        = {Context Repositories: Version-Controlled Memory for Agents},
  author       = {{Letta}},
  year         = {2026},
  howpublished = {Letta Blog},
  note         = {\url{https://www.letta.com/blog/context-repositories/}}
}

@article{kontonis2026memento,
  title={Memento: Teaching llms to manage their own context},
  author={Kontonis, Vasilis and Zeng, Yuchen and Garg, Shivam and Chen, Lingjiao and Tang, Hao and Wang, Ziyan and Awadallah, Ahmed and Horvitz, Eric and Langford, John and Papailiopoulos, Dimitris},
  journal={arXiv preprint arXiv:2604.09852},
  year={2026}
}

@inproceedings{jimenez2024swebench,
  title     = {{SWE-bench}: Can Language Models Resolve Real-World GitHub Issues?},
  author    = {Jimenez, Carlos E. and Yang, John and Wettig, Alexander and
               Yao, Shunyu and Pei, Kexin and Press, Ofir and
               Narasimhan, Karthik R.},
  booktitle = {International Conference on Learning Representations},
  year      = {2024}
}

@article{zheng2025deepresearcher,
  title   = {{DeepResearcher}: Scaling Deep Research via Reinforcement
             Learning in Real-World Environments},
  author  = {Zheng, Yuxiang and Fu, Dayuan and Hu, Xiangkun and Cai, Xiaojie
             and Ye, Lyumanshan and Lu, Pengrui and Liu, Pengfei},
  journal = {arXiv preprint arXiv:2504.03160},
  year    = {2025}
}

@misc{wei2025browsecomp,
      title={BrowseComp: A Simple Yet Challenging Benchmark for Browsing Agents}, 
      author={Jason Wei and Zhiqing Sun and Spencer Papay and Scott McKinney and Jeffrey Han and Isa Fulford and Hyung Won Chung and Alex Tachard Passos and William Fedus and Amelia Glaese},
      year={2025},
      eprint={2504.12516},
      archivePrefix={arXiv},
      primaryClass={cs.CL},
      url={https://arxiv.org/abs/2504.12516}, 
}

@article{beam,
  title   = {Beyond a Million Tokens: Benchmarking and Enhancing Long-Term Memory in {LLMs}},
  author  = {Tavakoli, Mohammad and Salemi, Alireza and Ye, Carrie and Abdalla, Mohamed and Zamani, Hamed and Mitchell, J Ross},
  journal = {arXiv preprint arXiv:2510.27246},
  year    = {2025},
}

@misc{hindsight,
  title        = {Hindsight Is \#1 on {BEAM} --- the Benchmark That Tests Memory at 10{M} Tokens},
  author       = {Bartholomew, Ben},
  howpublished = {\url{https://hindsight.vectorize.io/blog/2026/04/02/beam-sota}},
  year         = {2026},
}

@misc{zep-research,
  title        = {Research: {Zep} Benchmark Results},
  author       = {{Zep}},
  howpublished = {\url{https://www.getzep.com/research/}},
  year         = {2026},
}

@article{mem0,
  title   = {{Mem0}: Building Production-Ready {AI} Agents with Scalable
             Long-Term Memory},
  author  = {Chhikara, Prateek and Khant, Dev and Aryan, Saket and
             Singh, Taranjeet and Yadav, Deshraj},
  journal = {arXiv preprint arXiv:2504.19413},
  year    = {2025}
}

@misc{mem0-eval,
  title        = {Memory Evaluation},
  author       = {{Mem0}},
  howpublished = {\url{https://docs.mem0.ai/core-concepts/memory-evaluation}},
  year         = {2026},
}

@misc{cognee-beam,
  title        = {cognee on {BEAM}: {SOTA} Results Without a
                  Benchmark-Specific Memory System},
  author = {Markovi{\'c}, Vasilije},
  howpublished = {\url{https://www.cognee.ai/blog/deep-dives/benchmarking-cognee-on-beam}},
  year         = {2026},
}

@misc{honcho,
  title        = {Honcho: Memory Infrastructure for Stateful Agents},
  author       = {{Plastic Labs}},
  howpublished = {\url{https://github.com/plastic-labs/honcho}},
  year         = {2026},
}

@misc{emergencemem,
  title        = {{SOTA} on {LongMemEval} with {RAG}},
  author       = {{Emergence AI}},
  howpublished = {\url{https://www.emergence.ai/blog/sota-on-longmemeval-with-rag}},
  year         = {2026},
}

@inproceedings{react,
  title     = {{ReAct}: Synergizing Reasoning and Acting in Language Models},
  author    = {Yao, Shunyu and Zhao, Jeffrey and Yu, Dian and Du, Nan and
               Shafran, Izhak and Narasimhan, Karthik and Cao, Yuan},
  booktitle = {International Conference on Learning Representations (ICLR)},
  year      = {2023}
}

@misc{ptc,
  title        = {Programmatic Tool Calling},
  author       = {{Anthropic}},
  howpublished = {\url{https://platform.claude.com/docs/en/agents-and-tools/tool-use/programmatic-tool-calling}},
  year         = {2025},
}

@misc{minimax-m3,
  title        = {{MiniMax} {M3}: Frontier Coding, {1M} Context, Native Multimodality --- All in One Model},
  author       = {{MiniMax}},
  howpublished = {\url{https://www.minimax.io/blog/minimax-m3}},
  year         = {2026},
}

@misc{harbor,
  title        = {Harbor: A Framework for Building and Running Agent
                  Evaluations at Scale},
  author       = {{Harbor Framework Team}},
  howpublished = {\url{https://github.com/laude-institute/harbor}},
  year         = {2026},
}

@inproceedings{cao2026remember,
  title={Remember me, refine me: A dynamic procedural memory framework for experience-driven agent evolution},
  author={Cao, Zouying and Deng, Jiaji and Yu, Li and Zhou, Weikang and Liu, Zhaoyang and Ding, Bolin and Zhao, Hai},
  booktitle={Findings of the Association for Computational Linguistics: ACL 2026},
  pages={16803--16822},
  year={2026}
}

@misc{codeact,
      title={Executable Code Actions Elicit Better LLM Agents}, 
      author={Xingyao Wang and Yangyi Chen and Lifan Yuan and Yizhe Zhang and Yunzhu Li and Hao Peng and Heng Ji},
      year={2024},
      eprint={2402.01030},
      archivePrefix={arXiv},
      primaryClass={cs.CL},
      url={https://arxiv.org/abs/2402.01030}, 
}

@misc{zep-locomo-blog,
  title        = {Lies, Damn Lies, and Statistics: Is Mem0 Really {SOTA} in Agent Memory?},
  author       = {{Zep}},
  year         = {2025},
  howpublished = {\url{https://blog.getzep.com/lies-damn-lies-statistics-is-mem0-really-sota-in-agent-memory/}},
}

@misc{mem0-zep-issue,
  title        = {Revisiting {Zep}'s 84\% {LoCoMo} Claim: Corrected Evaluation \& 58.44\% Accuracy},
  author       = {{Mem0}},
  year         = {2025},
  howpublished = {\url{https://github.com/getzep/zep-papers/issues/5}},
}

@misc{qwenpaw,
  title        = {{QwenPaw}},
  author       = {{Agentscope Team}},
  year         = {2026},
  howpublished = {\url{https://qwenpaw.agentscope.io/}},
}

@misc{mastra,
  title        = {Observational Memory: 95\% on {LongMemEval}},
  author       = {Barnes, Tyler},
  year         = {2026},
  month        = feb,
  howpublished          = {\url{https://mastra.ai/research/observational-memory}},
}

@misc{exabase,
  title        = {Exabase Reports State-of-the-Art Results on {BEAM} Memory Benchmark},
  author       = {{Exabase}},
  year         = {2026},
  month        = jul,
  day          = {28},
  howpublished          = {\url{https://www.hpcwire.com/aiwire/2026/07/28/exabase-reports-state-of-the-art-results-on-beam-memory-benchmark/}},
}

\clearpage
\appendix
\appendix
\section{Detailed Breakdowns of Benchmark Results}
\label{app:breakdown}

\subsection{Per-Question-Type Results on LongMemEval}
\label{app:lme-breakdown}

Table~\ref{tab:lme-breakdown} reports Scroll's per-question-type accuracy on
the \texttt{S} and \texttt{M} splits of LongMemEval under the protocol of
Section~\ref{sec:setup-protocol}. The degradation from \texttt{S} to \texttt{M}
is concentrated in question types that require aggregating evidence across many
sessions: multi-session accuracy drops from 88.0\% to 81.2\% and single-session
(preference) from 100.0\% to 83.3\%. With more irrelevant sessions, the
agent-written code reliably locates a single supporting session but often misses
part of the evidence.

\begin{table}[h]
  \centering
  \caption{Per-question-type accuracy (\%) of Scroll on LongMemEval$_S$ and
    LongMemEval$_M$ (backbone: Qwen3.8-Max). Question types follow the
    benchmark taxonomy~\citep{wu2025longmemeval}.}
  \label{tab:lme-breakdown}
  \small
  \begin{tabular}{lcc}
    \toprule
    Question type               & LongMemEval$_S$ & LongMemEval$_M$ \\
    \midrule
    Single-session (user)       & 98.6            & 100.0           \\
    Single-session (assistant)  & 100.0           & 100.0           \\
    Single-session (preference) & 100.0           & 83.3            \\
    Multi-session               & 88.0            & 81.2            \\
    Temporal reasoning          & 94.0            & 93.6            \\
    Knowledge update            & 98.7            & 87.2            \\
    \midrule
    Overall                     & 94.8            & 89.6            \\
    \bottomrule
  \end{tabular}
\end{table}

\subsection{Per-Category Results on BEAM}
\label{app:beam-category}
Table~\ref{tab:beam-category} breaks BEAM$_{10M}$ down by memory ability. We compare against Mem0, Hindsight, and Exabase~M-1, the baselines for which per-category BEAM$_{10M}$ results are publicly available as of August~15, 2026~\citep{mem0-eval,hindsight,exabase}; Cognee and Honcho report only overall scores. Mem0, as a representative of the ingestion-heavy, fixed-pipeline paradigm, makes the contrast with Scroll's query-time approach most visible at the category level.

\begin{table}[h]
  \centering
  \caption{Per-category judge scores on BEAM$_{10M}$. Categories follow the
    benchmark's ten memory abilities~\citep{beam}. Scroll uses Qwen3.8-Max with
    thinking on; baseline breakdowns are taken from their published evaluations and
    use different configurations, so cross-system comparison is indicative rather than
    controlled.}
  \label{tab:beam-category}
  \small
  \setlength{\tabcolsep}{4pt}
  \begin{tabular}{lcccc}
    \toprule
    Category                 & Mem0          & Hindsight     & Exabase M-1   & Scroll (ours) \\
    \midrule
    Information extraction   & 56.2          & 51.2          & 66.3          & \textbf{75.0} \\
    Multi-session reasoning  & \textbf{26.1} & 16.6          & 9.6           & 21.9          \\
    Event ordering           & 20.2          & 61.6          & \textbf{67.5} & 64.1          \\
    Temporal reasoning       & 16.3          & 41.3          & \textbf{58.8} & 47.5          \\
    Knowledge update         & 75.0          & 65.0          & 45.0          & \textbf{92.5} \\
    Contradiction resolution & 32.5          & 56.9          & 58.8          & \textbf{88.1} \\
    Summarization            & 46.9          & 78.2          & \textbf{91.9} & 70.5          \\
    Preference following     & 90.4          & \textbf{97.5} & 95.0          & 89.1          \\
    Instruction following    & 82.5          & 92.5          & \textbf{97.5} & \textbf{97.5} \\
    Abstention               & 40.0          & 80.0          & \textbf{90.0} & 85.0          \\
    \midrule
    Overall                  & 48.6          & 64.1          & 68.0          & \textbf{73.1} \\
    \bottomrule
  \end{tabular}
\end{table}

Scroll leads by the widest margins where the answer hinges on a few exact records that must be located in the raw history and then ordered or reconciled: knowledge update (92.5 vs.\ 45.0--75.0), contradiction resolution (88.1 vs.\ 32.5--58.8), and information extraction (75.0 vs.\ 51.2--66.3). These categories punish write-time compression: resolving an update or a contradiction needs both sides of the value timeline, in order, with provenance, whereas retrieval over an ingested store typically surfaces the current state of a fact without its ordered history. The degree varies by system (Mem0's add-only extraction preserves old facts and holds up on knowledge update at 75.0), but none matches recovering the evidence by address from the verbatim Event Log.

Conversely, Scroll underperforms the strongest baselines where the graded artifact is itself a condensed view over many records: summarization (70.5 vs.\ 91.9 for Exabase M-1), preference following (89.1 vs.\ 97.5 for Hindsight), and temporal reasoning (47.5 vs.\ 58.8 for Exabase M-1). Ingestion-heavy pipelines build digests and preference profiles at write time, so the condensed view already exists when the question arrives; Scroll must reconstruct it from raw events per query, and its residual failures there are errors of query formulation rather than retrieval (Appendix~\ref{app:case-studies}).

Multi-session reasoning remains the weakest category for every system (9.6--26.1); Scroll's misses stem from over-precise filters that undercount the evidence set rather than from unreachable records. This profile is consistent across our repeated runs: contradiction resolution and knowledge update are among Scroll's strongest categories in every run, multi-session reasoning and summarization among its weakest.

\section{Additional LOCA Results}
\label{app:loca-full}

\paragraph{Comparison with published results.}
Table~\ref{tab:loca-published} places Scroll alongside the best publicly
reported results from the LOCA paper~\citep{zeng2026loca} and its leaderboard.
These systems use different backbone models, so the comparison is
system-level rather than controlled.

\begin{table}[h]
  \centering
  \caption{System-level comparison on LOCA. Results for prior systems are
    taken from the LOCA paper~\citep{zeng2026loca} or its public leaderboard and use
    different backbone models; ``--'' denotes no publicly reported result.}
  \label{tab:loca-published}
  \small
  \begin{tabular}{lcc}
    \toprule
    System                                       & 128K          & 256K          \\
    \midrule
    GPT-5.2-Medium + ReAct~\citep{react}         & 38.7          & 21.3          \\
    GPT-5.2-Medium + PTC~\citep{ptc}             & 49.3          & --            \\
    Claude-4.5-Opus + ReAct~\citep{zeng2026loca} & 34.0          & 14.7          \\
    MiniMax M3 + ReAct~\citep{minimax-m3}        & --            & 49.3          \\
    \midrule
    Scroll (Qwen3.8-Max)                         & \textbf{89.3} & \textbf{86.7} \\
    \bottomrule
  \end{tabular}
\end{table}

\section{Full Prompts and Rubrics}
\label{app:prompts}

Beyond the shared system prompt and context-management rules of
Section~3.2, each memory benchmark contributes one rubric, reproduced
below. For LOCA we use the benchmark's official task instructions
unmodified, adding only environment metadata (available APIs and
workspace paths).

\newenvironment{rubricbox}[1]{%
  \begin{tcolorbox}[colback=codebg, colframe=codeframe, boxrule=0.6pt,
      left=6pt, right=6pt, top=5pt, bottom=5pt, breakable]%
    \textbf{#1}\\[4pt]\ttfamily\footnotesize\raggedright\hbadness=10000
    }{\end{tcolorbox}}

\begin{rubricbox}{BEAM rubric}
  The user\textquotesingle{}s benchmark conversation history is stored in your durable history as rows with kind=\textquotesingle{}beam\_chat\_turn\textquotesingle{}. Rows are chronologically ordered by seq; each session has a distinct session\_id; rows carry an ISO created\_at.

  \medskip

  Recall it with the recall\_history\_python tool: pass a Python cell that uses the pre-bound ms surface. Search with concise keyword or synonym queries, e.g. ms.search(QUERY, all\_agents=True, kind=\textquotesingle{}beam\_chat\_turn\textquotesingle{}, k=10). Uppercase OR passes through as a boolean operator; otherwise terms are AND-combined. Start with k=5 or k=10 and increase it only when the evidence is insufficient. If a question combines multiple named systems, efforts, or entities, search each one separately instead of putting every name into one AND query. Extract one directly relevant value for each part, preserve its unit, and only then calculate. Each hit already includes the full turn text together with its seq, session\_id, role, and metadata; do not call ms.expand merely to pair a user message with its assistant reply. If you must reread a returned turn, call ms.expand only with that hit\textquotesingle{}s exact seq neighbours (lo, hi). For a question about one source date or an inclusive date range, filter on the created\_at column (ISO-8601 text, lexically sortable) with ms.sql\_query, e.g. "SELECT seq, session\_id, role, created\_at, content FROM hist.conversation\_history WHERE kind=\textquotesingle{}beam\_chat\_turn\textquotesingle{} AND substr(created\_at,1,10) BETWEEN \textquotesingle{}2024-07-01\textquotesingle{} AND \textquotesingle{}2024-07-31\textquotesingle{} ORDER BY seq LIMIT 50". For elapsed calendar days between two dates, use ms.days\_between(d1, d2).

  \medskip

  Treat role=\textquotesingle{}user\textquotesingle{} rows as evidence of the user\textquotesingle{}s facts, actions, and preferences; an assistant suggestion is not evidence that the user adopted it. Preserve exact numbers, units, and version labels from the most directly relevant user evidence; do not replace them with illustrative values. Do not add repeated historical mentions as separate quantities unless the question explicitly asks for that. When a fact changed, use the latest user evidence only after confirming that the rows describe the same project and the same fact, not merely similarly named work. Base your answer only on recalled conversation evidence. Do not use information from other history kinds. Follow any output-count or formatting constraint in the question exactly. If the requested fact is absent, say so clearly.

  \medskip
  Grounding rule (strict): every concrete claim --- each fact, number, name, date, quantity, or event --- must come verbatim in meaning from a turn you retrieved. Never invent specifics to make an answer sound complete.

  \medskip
  Decide between answering and saying "not enough information" by what you actually retrieved, not by how hard you searched: if a turn directly states the asked-for fact, give it; if none does, say there is not enough information in the conversation. A turn on a merely related topic is not the fact, and "not enough information" is itself a correct, expected answer.

  \medskip

  Always finish with submit\_answer and a non-empty, natural-language answer. Once more searching stops improving your answer, commit it rather than continuing until you run out. Never end without one.

\end{rubricbox}

\begin{rubricbox}{LongMemEval rubric}
  The user\textquotesingle{}s benchmark conversation history is stored in your durable history as rows with kind=\textquotesingle{}context\_msg\textquotesingle{} (user turns) or kind=\textquotesingle{}model\_turn\textquotesingle{} (assistant turns). Every turn\textquotesingle{}s content opens with a {[}Session N | YYYY-MM-DD{]} role: tag. Rows are chronologically ordered by seq; each session has a distinct session\_id; rows carry an ISO created\_at.

  \medskip

  Recall it with the recall\_history\_python tool: pass a Python cell that uses the pre-bound ms surface. Search with concise keyword or synonym queries, e.g. ms.search(QUERY, all\_agents=True, kind=\textquotesingle{}context\_msg\textquotesingle{}, k=10). Uppercase OR passes through as a boolean operator; otherwise terms are AND-combined. Start with k=5 or k=10 and increase it only when the evidence is insufficient. If a question combines multiple named systems, efforts, or entities, search each one separately instead of putting every name into one AND query. Extract one directly relevant value for each part, preserve its unit, and only then calculate. Each hit already includes the full turn text together with its seq, session\_id, role, and metadata; do not call ms.expand merely to pair a user message with its assistant reply. If you must reread a returned turn, call ms.expand only with that hit\textquotesingle{}s exact seq neighbours (lo, hi). For a question about one source date or an inclusive date range, filter on the created\_at column (ISO-8601 text, lexically sortable) with ms.sql\_query, e.g. "SELECT seq, session\_id, role, created\_at, content FROM hist.conversation\_history WHERE kind=\textquotesingle{}context\_msg\textquotesingle{} AND substr(created\_at,1,10) BETWEEN \textquotesingle{}2023-04-01\textquotesingle{} AND \textquotesingle{}2023-04-30\textquotesingle{} ORDER BY seq LIMIT 50". For elapsed calendar days between two dates, use ms.days\_between(d1, d2).

  \medskip

  Treat role=\textquotesingle{}user\textquotesingle{} rows as evidence of the user\textquotesingle{}s facts, actions, and preferences; an assistant suggestion is not evidence that the user adopted it. Preserve exact numbers, units, and version labels from the most directly relevant user evidence; do not replace them with illustrative values. Do not add repeated historical mentions as separate quantities unless the question explicitly asks for that. When a fact changed, use the latest user evidence only after confirming that the rows describe the same project and the same fact, not merely similarly named work. Base your answer only on recalled conversation evidence. Do not use information from other history kinds. Follow any output-count or formatting constraint in the question exactly. If the requested fact is absent, say so clearly.

  \medskip

  GROUNDING (strict): every concrete claim -- each fact, number, name, date, quantity, or event -- must come verbatim in meaning from a turn you retrieved. Never invent specifics to make an answer sound complete.

  \medskip

  Decide between answering and abstaining by what you actually retrieved, not by how hard you searched: if a turn directly states the asked-for fact, give it; if none does, abstain with the exact phrase "I don\textquotesingle{}t have that information from our conversations." A turn on a merely related topic is not the fact, and abstaining is itself a correct, expected answer when the fact is truly absent.

  \medskip

  Always finish with submit\_answer and a non-empty, natural-language answer. Once more searching stops improving your answer, commit it rather than continuing until you run out. Never end without one.
\end{rubricbox}

\section{Example Trajectories}
\label{app:case-studies}

This appendix reproduces four trajectories from the BEAM$_{10M}$ run reported in Table~\ref{tab:beam-category} (Scroll with Qwen3.8-Max, extended reasoning enabled): two successes from the categories where Scroll scores highest (knowledge update, 92.5; contradiction resolution, 88.1) and two failures from the categories where it trails the best published systems (preference following, 89.1 vs.\ 97.5 for Hindsight; summarization, 70.5 vs.\ 91.9 for Exabase M-1).

Each trajectory is shown in its logged JSON format (\texttt{task\_id}, \texttt{metrics}, \texttt{steps}), abridged as follows. 
Model-authored code
is moved from each step's \texttt{"source"} field into the referenced code
block and is otherwise unedited, including the model's own comments.
\texttt{"reasoning"} (the model's thinking before the action) and
\texttt{"observation"} strings are excerpted, with elisions marked
\texttt{[...]}; newlines inside strings are rendered literally
, and non-ASCII
punctuation is normalized. Gray \texttt{//} lines are our annotations, not
part of the log; the steps our trajectory analysis identifies as decisive for
the probe's outcome are flagged in bold color
(\trajgoodline{\texttt{//~>>}}~steps that drive the success,
\trajbadline{\texttt{//~>>}}~steps where the failure is decided).
\texttt{seq} denotes an event's immutable Event Log address;
\texttt{[Session n | date]} prefixes are part of the stored events.
Judge scores are those of the benchmark-provided LLM judge.

\subsection{Success (knowledge update): resolving a superseded value by Event Log order}
\label{app:traj-ku}

\begin{tcolorbox}[colback=codebg, colframe=codeframe, boxrule=0.6pt, left=6pt, right=6pt, top=5pt, bottom=5pt]
\textbf{Probe} (task \texttt{10M-1}, category \emph{knowledge update}; judge score 1.0 on all four rubric criteria):\\[3pt]
\emph{``How many tasks have I logged in Jira for the sprint on 2024-11-05,
and what is my sprint completion target percentage?''}
\end{tcolorbox}

The queried values are updated \emph{within} the relevant
session (Session 42; 14$\to$17 tasks, 85\%$\to$88\%), amid decoy sprints in Sessions 4,
22, and 47. The trajectory resolves the update by Event Log order: one
search, one SQL materialization, then in-kernel filters over the cached
variable.

\begin{lstlisting}[style=traj]
{"task_id": "beam/10M-1",
 "metrics": {"tokens_in": 51224, "tokens_out": 1471, "wall_time_s": 36.62, "step_count": 5},
 "steps": [
  (*@\trajgoodline{// >> Success: One coarse search pins the right session (42) among the}@*)
  (*@\trajgoodline{// >> decoy sprints (Sessions 4, 22, 47), and both competing values already surface}@*)
  (*@\trajgoodline{// >> in the hits}@*)
  {"step_index": 0,
   "reasoning": "I need to find the mentions of Jira tasks for the sprint on 2024-11-05, and the sprint completion target percentage. Let me search for these.",
   "action": {"tool": "recall_history_python",
              "args": {"source": (*@Code Block~\ref{lst:ku-c0}@*)}},
   "observation": "
      stdout:
      8349 42 [Session 42 | 2024-11-05] user: I'm trying to update the task count in Jira to reflect the new total of 17 tasks, and I want to make sure I'm doing it correctly to meet our sprint completion target of
      [...]
      8123 42 [Session 42 | 2024-11-05] user: I'm trying to plan my sprint and I've added 14 tasks for dynamic resizing, aiming for 85% sprint completion. I'm using Jira 9
      [...]
      8350 42 [Session 42 | 2024-11-05] assistant: Great! Let's update the task count in Jira to reflect the new total of 17 tasks and ensure you're on track to meet your sprint completion target of 88%.
      [...]
      565 4 [Session 4 | 2024-07-05] user: I've added 10 feasibility tasks to my Jira board, targeting 85% completion within t
      [...]
      4283 22 [Session 22 | 2024-08-05] user: I'm using Jira 9.5.0 to manage my project, and I've added 18 tasks for ingestion strateg
      [...]"},

  // The queried values changed within Session 42 -- materialize the whole session in
  // address order, bound to the kernel variable rows
  {"step_index": 1,
   "reasoning": "The search results show Session 42 (2024-11-05) which is the relevant session. [...] the value CHANGED over time. Let me look at the full session 42 to find the most recent values mentioned. [...]",
   "action": {"tool": "recall_history_python",
              "args": {"source": (*@Code Block~\ref{lst:ku-c1}@*)}},
   "observation": "
      stdout:
      seq=8047 role=user: [Session 42 | 2024-11-05] user: I'm working on enhancing the RAG system's context window management, specifically focusing on dynamic resizing based on query complexity, an
      [...]
      [output too long: 96806 chars printed, over the 32000-char limit -- the rest is hidden to protect your context window. Your variables persist, so re-run printing LESS: a count or list of seqs, snippet=True for a bounded triage view, or aggregate in a variable and print only the result -- not whole rows.]"},

  // The 32K observation cap truncated the print, but rows persists -- 
  // re-filter in the kernel with no Event Log re-access
  {"step_index": 2,
   "reasoning": "The result was too long. Let me print less - just the user turns and key snippets about task counts and sprint completion percentages from Session 42.",
   "action": {"tool": "recall_history_python",
              "args": {"source": (*@Code Block~\ref{lst:ku-c2}@*)}},
   "observation": "
      [...]
      seq=8047: [Session 42 | 2024-11-05] user: I'm working on enhancing the RAG system's context window management, specifically focus
      [...]"},

  (*@\trajgoodline{// >> Success: establishes intra-session order by seq -- 8123 (14 tasks,}@*)
  (*@\trajgoodline{// >> 85\%) precedes 8349 (17 tasks, 88\%) -- exactly the evidence needed to ensure}@*)
  (*@\trajgoodline{// >> latest information holds}@*)
  {"step_index": 3,
   "reasoning": "I found key evidence: Session 42 on 2024-11-05. [...] I need to check if there are any later mentions in session 42 that update these values further. [...]",
   "action": {"tool": "recall_history_python",
              "args": {"source": (*@Code Block~\ref{lst:ku-c3}@*)}},
   "observation": "
      [...]
      seq=8123: [Session 42 | 2024-11-05] user: I'm trying to plan my sprint and I've added 14 tasks for dynamic resizing, aiming for 85% sprint completion. I'm using Jira 9.6.0 to track my progress, but I'm not sure how to prioritize my tasks and allocate my time e
      [...]
      seq=8349: [Session 42 | 2024-11-05] user: I'm trying to update the task count in Jira to reflect the new total of 17 tasks, and I want to make sure I'm doing it correctly to meet our sprint completion target of 88%
      [...]"},

  // later address supersedes earlier: seq 8349 (17 tasks, 88%) follows seq 8123 (14
  // tasks, 85%); Session 47's sprint is ruled out by date.
  {"step_index": 4,
   "reasoning": "[...] 1. seq=8123: User says "I've added 14 tasks for dynamic resizing, aiming for 85% sprint completion" [...] The question asks specifically about "the sprint on 2024-11-05". The most recent mention within that sprint is seq=8349 with 17 tasks and 88% completion target. [...]",
   "action": {"tool": "submit_answer",
              "args": {"answer": "
      For the sprint on 2024-11-05, you logged **17 tasks** in Jira (updated from an earlier count of 14), and your sprint completion target percentage is **88%** (updated from the initial 85%)."}}}
 ]}
\end{lstlisting}

\begin{lstlisting}[style=scroll, caption={Knowledge update: full-text search; each hit returns its Event Log address (\texttt{seq}), session, and snippet.}, label={lst:ku-c0}]
hits = ms.search("sprint 2024-11-05", scope='task', k=20)
for h in hits:
    print(h["seq"], h["step_index"], h["snippet"][:200])
\end{lstlisting}

\begin{lstlisting}[style=scroll, caption={SQL materialization of Session 42 in address order, bound to the kernel variable \texttt{rows}.}, label={lst:ku-c1}]
rows = ms.sql_query(
    "SELECT seq, role, content FROM hist.conversation_history WHERE kind='conversation' AND step_index=42 ORDER BY msg_index"
)
for r in rows:
    # Print just the first 300 chars to get an overview
    print(f"seq={r['seq']} role={r['role']}: {r['content'][:300]}")
    print("---")
\end{lstlisting}

\begin{lstlisting}[style=scroll, caption={In-kernel filter of the cached \texttt{rows} (user turns only); no Event Log access.}, label={lst:ku-c2}]
# Print only user turns from session 42, just first 200 chars each
for r in rows:
    if r['role'] == 'user':
        print(f"seq={r['seq']}: {r['content'][:200]}")
        print("---")
\end{lstlisting}

\begin{lstlisting}[style=scroll, caption={Second in-kernel filter (user turns, \texttt{seq}${\geq}$8100), exposing the value progression.}, label={lst:ku-c3}]
# Look at user turns from seq 8100 onwards in session 42 to see progression
for r in rows:
    if r['role'] == 'user' and r['seq'] >= 8100:
        # Print seq and a snippet around "task" or "sprint"
        content = r['content']
        print(f"seq={r['seq']}: {content[:250]}")
        print("---")
\end{lstlisting}

\subsection{Success (contradiction resolution): retrieving both sides of a conflict and ordering them}
\label{app:traj-cr}

\begin{tcolorbox}[colback=codebg, colframe=codeframe, boxrule=0.6pt, left=6pt, right=6pt, top=5pt, bottom=5pt]
\textbf{Probe} (task \texttt{10M-10}, category \emph{contradiction
resolution; judge score 1.0 on all four rubric criteria}):\\[3pt]
\emph{``Has Pamela ever helped coordinate with vendors or saved setup time
during my events?''}
\end{tcolorbox}

The rubric requires stating that the record is contradictory, citing the
user's correction, citing the earlier claim, and resolving which is correct.
The trajectory operationalizes this as a retrieval program: one query per
side of the conflict, provenance checks on both, and an address-bounded
search that proves no later reversal.

\begin{lstlisting}[style=traj]
{"task_id": "beam/10M-10",
 "metrics": {"tokens_in": 104160, "tokens_out": 5758, "wall_time_s": 132.78, "step_count": 10},
 "steps": [
  // search the affirmative side of the conflict
  {"step_index": 0,
   "reasoning": "[...] I need to find all mentions of Pamela and vendor coordination/setup time, identify the conflict, and determine which statement is the latest or a correction. [...]",
   "action": {"tool": "recall_history_python",
              "args": {"source": (*@Code Block~\ref{lst:cr-c0}@*)}},
   "observation": "
      stdout:
      16900 73  ... Certainly! It sounds like Pamela took initiative to gather quick ratings from three vendors, which helped streamline the feedback process and saved
      [...]
      8434 36  ... 1. **Morning Setup**:
         - **Pamela**: Supervise the arrival and setup of vendors and decorators.

      [...]
      7254 31  ... Great to hear that Pamela is here to help! Given her experience and maturity, she can be a valuable asset in managing t
      [...]"},

  (*@\trajgoodline{// >> Success: the model authors an opposing, negation-term query for the}@*)
  (*@\trajgoodline{// >> 'against' side of the conflict; the user's correction surfaces at seq 8867}@*)
  {"step_index": 1,
   "reasoning": "The search results are showing that Pamela coordinated with vendors. Next I need to find the contradictory information. [...]",
   "action": {"tool": "recall_history_python",
              "args": {"source": (*@Code Block~\ref{lst:cr-c1}@*)}},
   "observation": "
      [...]
      8867 37 [Session 37 | 2023-07-15] user: How can I make sure Pamela is aware she didn't actually brief any entertainers or save me coordination time, as that
      [...]"},

  // materialize the correction exactly
  {"step_index": 2,
   "reasoning": "The search result at seq 8867 (Session 37) looks like it has key information [...]",
   "action": {"tool": "recall_history_python",
              "args": {"source": (*@Code Block~\ref{lst:cr-c2}@*)}},
   "observation": "
      stdout:
      8867 user
      [Session 37 | 2023-07-15] user: How can I make sure Pamela is aware she didn't actually brief any entertainers or save me coordination time, as that was a misunderstanding?
      ---"},

  // session-scoped SQL around the correction; the original delegation (seq 8675) 
  // sits in the same session
  {"step_index": 3,
   "reasoning": "[...] Now I need to check what was said earlier about Pamela coordinating with vendors or saving setup time. [...]",
   "action": {"tool": "recall_history_python",
              "args": {"source": (*@Code Block~\ref{lst:cr-c3}@*)}},
   "observation": "
      [...]
      seq 8675 [user]: [Session 37 | 2023-07-15] user: What's the most efficient way to brief the 5 entertainers, as Pamela had taken on this task earlier?
      [...]"},

  // expand the later, seemingly re-affirming mentions from Session 73
  {"step_index": 4,
   "reasoning": "[...] Let me look at the seq 16900 and 16902 hits from Session 73, which seem to discuss Pamela's vendor ratings and saved time. [...]",
   "action": {"tool": "recall_history_python",
              "args": {"source": (*@Code Block~\ref{lst:cr-c4}@*)}},
   "observation": "
      [...]
      seq 16902 [assistant]:
      [Session 73 | 2023-08-22] assistant: You're welcome! I'm glad that clarifying Pamela's actions helped you understand how her quick calls saved you time and provided immediate insights.
      [...]
      2. **Time Saved**: By handling these calls herself, Pamela saved you approximately 15 minutes of follow-up time.
      [...]"},

  // provenance check: those are assistant turns -- verify what the user actually
  // said nearby; the user's own later confirmation surfaces
  {"step_index": 5,
   "reasoning": "[...] this is assistant-generated text, not the user's own statement. I need to check whether the user in session 73 actually said something that led to this assistant response, or if this wa [...]",
   "action": {"tool": "recall_history_python",
              "args": {"source": (*@Code Block~\ref{lst:cr-c5}@*)}},
   "observation": "
      [...]
      seq 16899 [user]: [Session 73 | 2023-08-22] user: Can you help me understand why Pamela called 3 vendors for quick ratings, and how it saved me 15 minutes of follow-up?
      [...]
      seq 16901 [user]: [Session 73 | 2023-08-22] user: Yeah, that makes sense. Pamela's quick calls definitely saved me time. It's good to get those immediate insights without having to follow up myself. Thanks for clarifying!
      [...]"},

  // the reasoning separates the two claims: 'coordination time' during setup 
  // (corrected) vs. post-event 'follow-up time' (confirmed)
  {"step_index": 6,
   "reasoning": "[...] The correction in Session 37 is about Pamela NOT briefing entertainers or saving coordination time (a misunderstanding). [...] In Session 73, the user confirms Pamela called 3 vendors for quick ratings and saved 15 minutes of follow-up time. [...]",
   "action": {"tool": "recall_history_python",
              "args": {"source": (*@Code Block~\ref{lst:cr-c6}@*)}},
   "observation": "
      [...]
      seq 8868 [assistant]: [Session 37 | 2023-07-15] assistant: To clarify the misunderstanding with Pamela and ensure she is aware that she did not actually brief any entertainers or save coordination time, it's important to communicate this information
      [...]"},

  // retrieve the claim side verbatim (Sessions 31 and 36)
  {"step_index": 7,
   "reasoning": "[...] **Earlier claims (pre-correction):** Multiple sessions (31, 36) had assistant suggestions about Pamela supervising vendor arrival/setup, coordinating with vendors, etc. [...]",
   "action": {"tool": "recall_history_python",
              "args": {"source": (*@Code Block~\ref{lst:cr-c7}@*)}},
   "observation": "
      [...]
      seq 7496 [assistant]: [Session 31 | 2023-07-10] assistant: Delegating tasks like decor oversight to someone like Pamela can have a significant positive impact on the overall timeli
      [...]
      seq 8434 [assistant]: [Session 36 | 2023-07-15] assistant: That sounds like an excellent plan! By leveraging Pamela's skills and experience, you can ensure that the guest check-in
      [...]"},

  (*@\trajgoodline{// >> Success: the disconfirming query -- an address-bounded search proving}@*)
  (*@\trajgoodline{// >> nothing after Session 73 overturns the correction. }@*)
  (*@\trajgoodline{// >> Across this run, successful trajectories issue such a query before }@*)
  (*@\trajgoodline{// >> submitting; but failed ones never do}@*)
  {"step_index": 8,
   "reasoning": "[...] The later statement (Session 73) would normally win. But I need to check if there's a later correction after Session 73 as well. [...]",
   "action": {"tool": "recall_history_python",
              "args": {"source": (*@Code Block~\ref{lst:cr-c8}@*)}},
   "observation": "
      stdout:
      18077 77 [Session 77 | 2023-08-24] user: Yeah, it definitely caused some frustration. Those 5 guests w
      [...]"},

  // each rubric criterion maps to evidence retrieved explicitly
  {"step_index": 9,
   "reasoning": "[...] So the resolution is nuanced: The user corrected that Pamela did NOT brief entertainers or save coordination time during setup (that was a misunderstanding), but later confirmed Pamela DID help by calling vendors for quick ratings, saving follow-up time [...]",
   "action": {"tool": "submit_answer",
              "args": {"answer": "
      The conversation contains conflicting information about this, and here's how it resolves:

      **The initial claim:** In earlier sessions (around Sessions 31-36), plans were made for Pamela to coordinate with vendors, supervise vendor arrivals/setup, and brief entertainers -- with the expectation that this would save you coordination time during the event.

      **Your correction (Session 37, July 15, 2023):** You explicitly clarified that this was a misunderstanding -- Pamela did NOT actually brief any entertainers or save you coordination time during setup. You asked how to make sure Pamela was aware of this correction.

      **A later, narrower confirmation (Session 73, August 22, 2023):** You confirmed that Pamela did call 3 vendors for quick ratings after the event, which saved you about 15 minutes of follow-up time. You said: "Pamela's quick calls definitely saved me time."

      **Resolution:** The broad claim that Pamela coordinated vendors and saved setup time during your events was corrected by you as a misunderstanding (Session 37). However, Pamela did help in a more limited, post-event way -- gathering quick vendor ratings that saved you ~15 minutes of follow-up (Session 73). So she didn't save setup time or coordinate vendors during the event itself, but she did contribute afterward with vendor feedback calls."}}}
 ]}
\end{lstlisting}

\begin{lstlisting}[style=scroll, caption={Contradiction resolution: full-text search on the claim side.}, label={lst:cr-c0}]
hits = ms.search("Pamela vendor", scope='task', k=20)
for h in hits:
    print(h["seq"], h["step_index"], h["snippet"][:150])
\end{lstlisting}

\begin{lstlisting}[style=scroll, caption={Negation-term search for the correction side.}, label={lst:cr-c1}]
hits2 = ms.search("Pamela didn't OR wasn't OR never OR not", scope='task', k=20)
for h in hits2:
    print(h["seq"], h["step_index"], h["snippet"][:150])
\end{lstlisting}

\begin{lstlisting}[style=scroll, caption={Exact materialization of the correction (\texttt{ms.expand}).}, label={lst:cr-c2}]
rows = ms.expand([8867])
for r in rows:
    print(r["seq"], r["role"])
    print(r["content"][:2000])
    print("---")
\end{lstlisting}

\begin{lstlisting}[style=scroll, caption={Session-scoped SQL over Session 37.}, label={lst:cr-c3}]
# Look at the assistant response after seq 8867 and surrounding context in session 37
rows = ms.sql_query("SELECT seq, role, content FROM hist.conversation_history WHERE kind='conversation' AND step_index=37 ORDER BY seq LIMIT 30")
for r in rows:
    print(f"seq {r['seq']} [{r['role']}]: {r['content'][:200]}")
    print("---")
\end{lstlisting}

\begin{lstlisting}[style=scroll, caption={Expansion of the post-correction mentions (\texttt{seq} 16900, 16902).}, label={lst:cr-c4}]
# Check later references (after the correction at seq 8867) about Pamela and vendor coordination/setup time
rows = ms.expand([16900, 16902])
for r in rows:
    print(f"seq {r['seq']} [{r['role']}]:")
    print(r["content"][:600])
    print("---")
\end{lstlisting}

\begin{lstlisting}[style=scroll, caption={Provenance check---the user turns around \texttt{seq} 16900.}, label={lst:cr-c5}]
# Check what the user said in session 73 around seq 16900
rows = ms.sql_query("SELECT seq, role, content FROM hist.conversation_history WHERE kind='conversation' AND step_index=73 AND seq BETWEEN 16890 AND 16910 ORDER BY seq")
for r in rows:
    print(f"seq {r['seq']} [{r['role']}]: {r['content'][:300]}")
    print("---")
\end{lstlisting}

\begin{lstlisting}[style=scroll, caption={The assistant's uptake of the correction (\texttt{seq} 8868).}, label={lst:cr-c6}]
# Check the assistant response after the user's correction at seq 8867
rows = ms.sql_query("SELECT seq, role, content FROM hist.conversation_history WHERE kind='conversation' AND step_index=37 AND seq BETWEEN 8867 AND 8870 ORDER BY seq")
for r in rows:
    print(f"seq {r['seq']} [{r['role']}]: {r['content'][:500]}")
    print("---")
\end{lstlisting}

\begin{lstlisting}[style=scroll, caption={The original claim side (Sessions 31 and 36).}, label={lst:cr-c7}]
# Let me also check the earlier sessions (31, 36) where Pamela was assigned vendor coordination roles, to get the full picture of the "claim" side
rows = ms.expand([7496, 8434])
for r in rows:
    print(f"seq {r['seq']} [{r['role']}]: {r['content'][:400]}")
    print("---")
\end{lstlisting}

\begin{lstlisting}[style=scroll, caption={\texttt{seq\_range}-bounded search over the remainder of the history, proving no later reversal.}, label={lst:cr-c8}]
# Check if there are any later corrections about Pamela after session 73
hits3 = ms.search("Pamela misunderstanding OR correction OR didn't OR wasn't", scope='task', k=15, seq_range=(16900, 23716))
for h in hits3:
    print(h["seq"], h["step_index"], h["snippet"][:200])
\end{lstlisting}

\subsection{Failure (preference following): competent retrieval on the wrong axis}
\label{app:traj-pf}

\begin{tcolorbox}[colback=codebg, colframe=codeframe, boxrule=0.6pt, left=6pt, right=6pt, top=5pt, bottom=5pt]
\textbf{Probe} (task \texttt{10M-8}, category \emph{preference following}; judge score 0.0 on all three rubric criteria):\\[3pt]
\emph{``I'm planning the route for our upcoming trip. How would you suggest
I map it out?''}
\end{tcolorbox}

All three
criteria concern one stated preference: avoiding tolls. The retrieval
patterns that succeed elsewhere all execute correctly here---the failure is
upstream, in query formulation: the model frames the probe as \emph{which
tool} the user prefers and never queries the \emph{route-constraint} axis.
The graded preference is abundant in the Event Log, in the user's own words
(seq 805: ``which route would you say is better for someone who wants to
avoid tolls as much as possible?''; seq 807; Session 3, seq 617--631, where
the hybrid route is adopted because it saves \$15 in tolls).

\begin{lstlisting}[style=traj]
{"task_id": "beam/10M-8",
 "metrics": {"tokens_in": 82830, "tokens_out": 2471, "wall_time_s": 69.19, "step_count": 7},
 "steps": [
  (*@\trajbadline{// >> Failure: the outcome is decided here, before any retrieval runs --}@*)
  (*@\trajbadline{// >> the probe is framed as 'which mapping tool', all six queries name tools, and}@*)
  (*@\trajbadline{// >> none of the trajectory's 14 queries contains toll, fee, avoid, or cost, }@*)
  (*@\trajbadline{// >> which is the axis all three rubric criteria grade}@*)
  {"step_index": 0,
   "reasoning": "[...] I need to find a preference they established earlier -- a chosen tool, version, style, or constraint. [...]",
   "action": {"tool": "recall_history_python",
              "args": {"source": (*@Code Block~\ref{lst:pf-c0}@*)}},
   "observation": "
      stdout:
      10016 43  ... Drag and drop the route lines to adjust the path according to the detour.
      [...]
      8148 34  ... Compare the routes to see which one avoids tolls.
      [...]
      662 3  ... Sync with Google Maps
      While you cannot directly sync Google Maps with GasBuddy, you can use both apps together
      [...]"},

  (*@\trajbadline{// >> Failure: the toll preference brushes past in the previous step's hits}@*)
  (*@\trajbadline{// >> (seq 8148, 'avoids tolls') and goes unpursued; the model chases the Maps.me }@*)
  (*@\trajbadline{// >> tool candidate instead}@*)
  {"step_index": 1,
   "reasoning": "[...] Let me also check seq 16614 which mentions Maps.me as a "Current Tool". [...]",
   "action": {"tool": "recall_history_python",
              "args": {"source": (*@Code Block~\ref{lst:pf-c1}@*)}},
   "observation": "
      [...]
      16614 [Session 75 | 2023-04-20] assistant: Improving your detour accuracy and saving time with navigation tools like Maps.me is a great achievement. Here are some other n
      [...]"},

  // sound provenance discipline: the Maps.me hit is discarded as assistant-authored;
  // five preference-phrase queries follow -- still tool-framed
  {"step_index": 2,
   "reasoning": "[...] that's an assistant response. I need to find where the user explicitly states their preference for a mapping tool. [...]",
   "action": {"tool": "recall_history_python",
              "args": {"source": (*@Code Block~\ref{lst:pf-c2}@*)}},
   "observation": "
      [...]
      2349 10  ... That sounds like a solid plan! I'll make sure to use Google Maps and Roadtrippers to map out the hybrid route and share it with Chris. I'll also download off
      [...]
      17175 77 [Session 77 | 2023-04-22] user: That sounds great! I'll definitely use mapping tools like Google Maps and Waze to plan my routes and find good stopping points.
      [...]"},

  // recency check over the final sessions: 
  // no later preference supersedes the finding
  {"step_index": 3,
   "reasoning": "[...] Let me look more closely at the most recent user turns about mapping, especially in later sessions (82-100), [...]",
   "action": {"tool": "recall_history_python",
              "args": {"source": (*@Code Block~\ref{lst:pf-c3}@*)}},
   "observation": "
      (no output)"},

  // user-authored evidence for the tool preference accumulates (Sessions 1, 10, 13)
  {"step_index": 4,
   "reasoning": "[...] the user has consistently used **Roadtrippers** and **Google Maps** together. [...]",
   "action": {"tool": "recall_history_python",
              "args": {"source": (*@Code Block~\ref{lst:pf-c4}@*)}},
   "observation": "
      [...]
      2335 10  ... That sounds perfect! I'll use Google Maps and Roadtrippers to map out the hybrid route and share it with Chris. We'll definitely need to download offline maps for a
      [...]
      25 [Session 1 | 2022-05-01] user: How does the Roadtrippers app work, and can it really help me visualize a route that covers at least 5 national parks?
      [...]"},

  // scope the upcoming trip (Session 100): 
  // a 1,200-mile northern route, four landmarks
  {"step_index": 5,
   "reasoning": "[...] check if there's anything in the most recent sessions (S100, seq 22221-22560) about a specific route they're planning for the "upcoming trip [...]",
   "action": {"tool": "recall_history_python",
              "args": {"source": (*@Code Block~\ref{lst:pf-c5}@*)}},
   "observation": "
      [...]
      22221 user [Session 100 | 2023-05-31] user: What's the best way to plan a 1,200-mile northern route within 9 months, aiming for 4 key landmarks, like I've targeted?
      [...]"},

  // The answer is grounded and well-structured, but without toll information. 
  // Judge: 0.0 on all three criteria.
  {"step_index": 6,
   "reasoning": "[...] The most recent and repeated preference is **Google Maps and Roadtrippers** used together. The user consistently says they'll use both to map out routes. [...]",
   "action": {"tool": "submit_answer",
              "args": {"answer": "
      Based on our past conversations, you've consistently used **Google Maps and Roadtrippers** together to map out your routes -- you mentioned this combination multiple times, including when planning the hybrid route with Chris and when scouting scenic areas. I'd suggest sticking with that approach for your upcoming 1,200-mile northern route to your 4 key landmarks:
      [...]
      1. **Roadtrippers first** -- Plot your 4 landmarks as waypoints and use it to discover scenic stops, detours, and points of interest along the northern route. It's great for visualizing the overall tri
      [...]
      This two-tool combo has worked well for you across multiple trips, so it should serve you well for this one too."}}}
 ]}
\end{lstlisting}

\begin{lstlisting}[style=scroll, caption={Preference following: multi-query sweep, de-duplicated by address; all six queries name tools.}, label={lst:pf-c0}]
# Search for mapping tool preferences in the conversation
hits = []
for q in ["map route", "Google Maps", "Roadtrippers", "mapping tool", "app route", "map out"]:
    hits += ms.search(q, scope='task', k=10)

# Print unique hits with snippets for triage
seen = set()
for h in hits:
    if h["seq"] not in seen:
        seen.add(h["seq"])
        print(h["seq"], h["step_index"], h.get("snippet", "")[:150])
        print("---")
\end{lstlisting}

\begin{lstlisting}[style=scroll, caption={Expansion of the Maps.me candidate (\texttt{seq} 16614).}, label={lst:pf-c1}]
# Let me look at the user's own turns about mapping tools, especially from later sessions
# First check seq 16614 which mentions Maps.me as "Current Tool"
rows = ms.expand([16614])
for r in rows:
    print(r["seq"], r["content"][:500])
    print("---")
\end{lstlisting}

\begin{lstlisting}[style=scroll, caption={Preference-phrase sweep plus a Roadtrippers query---still tool-framed.}, label={lst:pf-c2}]
# Search for user turns specifically about their preferred mapping tool
hits2 = []
for q in ["I'll use", "I prefer", "my favorite", "I've been using", "I use"]:
    hits2 += ms.search(f"{q} map", scope='task', k=10)

# Also search for Roadtrippers specifically in user turns
hits3 = ms.search("Roadtrippers route plan", scope='task', k=15)

seen2 = set()
for h in hits2 + hits3:
    if h["seq"] not in seen2:
        seen2.add(h["seq"])
        print(h["seq"], h["step_index"], h.get("snippet", "")[:160])
        print("---")
\end{lstlisting}

\begin{lstlisting}[style=scroll, caption={\texttt{seq\_range}-bounded recency check over the final sessions.}, label={lst:pf-c3}]
# Look at user turns in later sessions about mapping tools
hits4 = ms.search("map route plan", seq_range=(18323, 22716), scope='task', k=20)
for h in hits4:
    if h.get("role") == "user":
        print(h["seq"], h["step_index"], h.get("snippet", "")[:180])
        print("---")
\end{lstlisting}

\begin{lstlisting}[style=scroll, caption={User-turn search plus session-scoped SQL over Session 1.}, label={lst:pf-c4}]
# Check user turns specifically mentioning their chosen mapping approach
# Focus on user turns that state what they'll use or prefer
hits5 = ms.search("Google Maps Roadtrippers map out", scope='task', k=15)
for h in hits5:
    if h.get("role") == "user":
        print(h["seq"], h["step_index"], h.get("snippet", "")[:200])
        print("---")

# Also check the earliest session where Roadtrippers was introduced
print("\n=== Session 1 user turns about Roadtrippers ===")
rows_s1 = ms.sql_query("SELECT seq, role, content FROM hist.conversation_history WHERE kind='conversation' AND step_index=1 AND role='user' AND content LIKE '%Roadtrippers%' ORDER BY msg_index LIMIT 5")
for r in rows_s1:
    print(r["seq"], r["content"][:300])
    print("---")
\end{lstlisting}

\begin{lstlisting}[style=scroll, caption={Reading the opening of Session 100 to scope the upcoming trip.}, label={lst:pf-c5}]
# Check the most recent session (S100) for context about the upcoming trip
rows_s100 = ms.sql_query("SELECT seq, role, content FROM hist.conversation_history WHERE kind='conversation' AND step_index=100 ORDER BY msg_index LIMIT 10")
for r in rows_s100:
    print(r["seq"], r["role"], r["content"][:250])
    print("---")
\end{lstlisting}

\subsection{Failure (summarization): positional sampling misses mid-session evidence}
\label{app:traj-sum}

\begin{tcolorbox}[colback=codebg, colframe=codeframe, boxrule=0.6pt, left=6pt, right=6pt, top=5pt, bottom=5pt]
\textbf{Probe} (task \texttt{10M-8}, category \emph{summarization}; judge
score 0.42---two of six rubric criteria fully satisfied, one partially):\\[3pt]
\emph{``Can you summarize how my travel decisions and habits evolved from
April 8, 2023 to April 25, 2023 and how they influenced my overall experience
and personal growth?''}
\end{tcolorbox}

Coverage at session granularity is complete: cells
0--6 page and merge all 4,448 events in the date range (Sessions 61--80,
verified gap-free). The score is then decided by sampling choices within that
coverage: the trajectory reads long mixed-topic sessions \emph{positionally}
(head and tail) rather than \emph{thematically}, and the graded mid-session
facts---the 3-hour driving cap with its fatigue reduction (Session 77, seq
17171--17207) and the sleep and hydration goals (Session 66, seq
14809--14817)---never enter an observation.

\begin{lstlisting}[style=traj]
{"task_id": "beam/10M-8",
 "metrics": {"tokens_in": 197973, "tokens_out": 8846, "wall_time_s": 221.2, "step_count": 14},
 "steps": [
  // date-range SQL over event metadata; both the 1000-row cap and the 32K observation
  // cap bite -- the model pivots to paging
  {"step_index": 0,
   "reasoning": "[...] First, I need to find the conversations within this date range. I'll try using SQL to filter by date.",
   "action": {"tool": "recall_history_python",
              "args": {"source": (*@Code Block~\ref{lst:sum-c0}@*)}},
   "observation": "
      stdout:
      1000
      13611 61 2023-04-08 user [Session 61 | 2023-04-08] user: What's the best way to spend our last day in Los Angeles bef
      [...]
      [note] a sql_query hit the 1000-row cap -- matching rows beyond the first 1000 were NOT returned; narrow with WHERE or page with LIMIT/OFFS
      [...]"},

  // aggregate per session in the kernel: page 1 reaches only 2023-04-13
  {"step_index": 1,
   "reasoning": "The result got cut off at 1000 rows. Need to paginate. [...]",
   "action": {"tool": "recall_history_python",
              "args": {"source": (*@Code Block~\ref{lst:sum-c1}@*)}},
   "observation": "
      stdout:
      1000 13611 14610 2023-04-13
      61 ['2023-04-08'] 188
      62 ['2023-04-09'] 222
      63 ['2023-04-10'] 196
      64 ['2023-04-11'] 190
      65 ['2023-04-12'] 194
      66 ['2023-04-13'] 10"},

  // page the remainder; steps 3-5 rerun this query changing only the OFFSET
  {"step_index": 2,
   "reasoning": "[...] I need to fetch the remaining period (April 13-25). [...]",
   "action": {"tool": "recall_history_python",
              "args": {"source": (*@Code Block~\ref{lst:sum-c2}@*)}},
   "observation": "
      stdout:
      1000
      66 ['2023-04-13'] 268
      67 ['2023-04-14'] 200
      68 ['2023-04-15'] 198
      69 ['2023-04-15'] 196
      70 ['2023-04-15'] 138
      15600 2023-04-15"},

  {"step_index": 3,
   "reasoning": "I need to continue paging through the data and keep collecting the remaining pages until I reach April 25th.",
   "action": {"tool": "recall_history_python",
              "args": {"source": (*@Code Block~\ref{lst:sum-c2} (OFFSET 1000)@*)}},
   "observation": "
      stdout:
      1000 16600 2023-04-20
      [...]"},

  {"step_index": 4,
   "reasoning": "On track -- continuing the paging to cover April 20-25.",
   "action": {"tool": "recall_history_python",
              "args": {"source": (*@Code Block~\ref{lst:sum-c2} (OFFSET 2000)@*)}},
   "observation": "
      stdout:
      1000 17600 2023-04-24
      [...]"},

  {"step_index": 5,
   "reasoning": "[...] Let me fetch the last page (the remainder from April 24th onwards) to complete the coverage, [...]",
   "action": {"tool": "recall_history_python",
              "args": {"source": (*@Code Block~\ref{lst:sum-c2} (OFFSET 3000)@*)}},
   "observation": "
      stdout:
      458
      79 ['2023-04-24'] 178
      80 ['2023-04-25'] 280"},

  // merge and de-duplicate by seq in the kernel: 4,448 events, Sessions 61-80, verified
  // gap-free -- session-level coverage is complete. Note Session 77's size: 470 events
  {"step_index": 6,
   "reasoning": "[...] Next I need to merge everything into one complete dataset (deduplication), [...]",
   "action": {"tool": "recall_history_python",
              "args": {"source": (*@Code Block~\ref{lst:sum-c6}@*)}},
   "observation": "
      [...]
      4448 13611 2023-04-08 18058 2023-04-25
      [...]
      61 ['2023-04-08'] 188 13611 13798
      [...]
      77 ['2023-04-22'] 470 16909 17378
      [...]
      80 ['2023-04-25'] 280 17779 18058"},

  (*@\trajbadline{// >> Failure: the first of the two sampling choices that decide the score}@*)
  (*@\trajbadline{// >> -- positional reading (first 3 + last 2 user turns per session);}@*)
  (*@\trajbadline{// >> head-and-tail reads of 200-470-row mixed-topic sessions systematically}@*)
  (*@\trajbadline{// >> skip the middle, where habit threads sit}@*)
  {"step_index": 7,
   "reasoning": "[...] that's about 2200 user turns, way too much. Better approach: for each session, print the first few and last few user turns.",
   "action": {"tool": "recall_history_python",
              "args": {"source": (*@Code Block~\ref{lst:sum-c7}@*)}},
   "observation": "
      [...]
      === Session 61 (2023-04-08) -- 94 user turns ===
      [...]
      13611: [S61 | 2023-04-08] U: What's the best way to spend our last day in Los Angeles before heading home, considering we're starting the final stretch of our 2,500-mi
      [...]"},

  (*@\trajbadline{// >> Failure: the deep read stays head-only (first 6 user turns of 7 sessions); }@*)
  (*@\trajbadline{// >> the graded driving-limit arc sits ~100 user turns into Session 77}@*)
  (*@\trajbadline{// >> (seq 17171-17207) and is never reached -- even though this read surfaces 'I'm}@*)
  (*@\trajbadline{// >> trying to manage my travel fatigue' (Session 63) as a lead}@*)
  {"step_index": 8,
   "reasoning": "[...] To grasp the main components, let me try sampling a few turns from each session in more detail. [...]",
   "action": {"tool": "recall_history_python",
              "args": {"source": (*@Code Block~\ref{lst:sum-c8}@*)}},
   "observation": "
      [...]
      ########## SESSION 61 (2023-04-08) ##########
      [...]
      --- seq 13611 ---
      [Session 61 | 2023-04-08] user: What's the best way to spend our last day in Los Angeles before heading home, considering we're starting the final stretch of ou
      [...]"},

  (*@\trajbadline{// >> Failure: the second sampling choice -- the trajectory's only thematic}@*)
  (*@\trajbadline{// >> search covers growth vocabulary exclusively; the probe asks about }@*)
  (*@\trajbadline{// >> 'decisions AND habits', but no query ever names driving hours, fatigue, }@*)
  (*@\trajbadline{// >> sleep, or hydration (the graded habit dimensions)}@*)
  {"step_index": 9,
   "reasoning": "[...] Let me search for growth-related turns and key decisions within a few sessions. [...]",
   "action": {"tool": "recall_history_python",
              "args": {"source": (*@Code Block~\ref{lst:sum-c9}@*)}},
   "observation": "
      [...]
      15489  ... I feel like I learned a lot about myself and our relationship during that hike.
      [...]"},

  // verbatim grounding of the growth quantifications (40% patience, 35% flexibility, 
  // 82% self-rating) -- these earn the fully satisfied criteria
  {"step_index": 10,
   "reasoning": "[...] Let me expand a few of the key seqs to get verbatim details. [...]",
   "action": {"tool": "recall_history_python",
              "args": {"source": (*@Code Block~\ref{lst:sum-c10}@*)}},
   "observation": "
      [...]
      === seq 15489 | role=user ===
      [Session 70 | 2023-04-15] user: Yeah, the Grand Canyon hike was definitely a standout moment. It pushed us both physically and mentally, and seeing the canyon was breathtaking. It made me realize how much I can accomplish with determination and teamwork. Plus, it gave me a new perspective on life and made me appreciate nature more. I feel like I learned a lot about myself and our relationship during that hike.


      [...]"},

  // budget details land ($70 Pittsburgh delay, $250 food overspend); the 
  // sleep/hydration cluster (seq 14809-14817, mid-Session 66) is never touched
  {"step_index": 11,
   "reasoning": "[...] let me gather a few more specific points: budget details ($250 food overspend, Pittsburgh delay $70), habit changes (session 72, screen time), [...]",
   "action": {"tool": "recall_history_python",
              "args": {"source": (*@Code Block~\ref{lst:sum-c11}@*)}},
   "observation": "
      [...]
      HIT 17368  ... In this case, the total overspend on food is $250, with $60 coming from a single diner visit in LA. Here's a structured approach to
      [...]"},

  // final spot checks; the model declares coverage sufficient
  {"step_index": 12,
   "reasoning": "[...] Actually, I have enough material. [...]",
   "action": {"tool": "recall_history_python",
              "args": {"source": (*@Code Block~\ref{lst:sum-c12}@*)}},
   "observation": "
      [...]
      [Session 79 | 2023-04-24] assistant: Limiting trips to 2 stops max played a significant role in my overall personal growth, particularly in handling unexpected detours and fee
      [...]"},

  // judge: 0.42 -- two of six criteria fully satisfied, one partial. Everything the
  // rubric graded was inside the materialized sessions; the misses (3-hour driving 
  // cap, Session 77; 8-hour sleep and hydration goals, Session 66) fell outside the 
  // sampled positions
  {"step_index": 13,
   "reasoning": "[...] Let me do a final coverage check against what the question is ask [...]",
   "action": {"tool": "submit_answer",
              "args": {"answer": "
      **The arc, April 8-25, 2023 (Sessions 61-80):** Your travel decisions evolved through three clear phases: (1) **wrap-up logistics** in Los Angeles (Apr 8-10), (2) **deliberate rest and the journey home** (Apr 13-16), and (3) **post-trip processing, sharing, and lesson-building** back in New Jeffreytow
      [...]
      You **quantified your growth**: 40% more patience (later framed as a 40% increase in confidence and adaptability), 35% more flexibility after the 5 unplanned detours, and an overall trip-growth self-rating of 82%.

      [...]
      By April 25 you had distilled **4 key trip lessons** -- "Flexibility is Key" and "Embrace Spontaneity" among them -- and begun journaling to sustain the gains. Your unplanned stops reshaped your future travel philosophy: looser itineraries with buffer time, openness to local recommendations, and fewer, deeper stops rather than rigid day-by-day plans."}}}
 ]}
\end{lstlisting}

\begin{lstlisting}[style=scroll, caption={Summarization: date-range SQL over event metadata (2023-04-08 to 2023-04-25).}, label={lst:sum-c0}]
rows = ms.sql_query(
    "SELECT seq, step_index, json_extract(metadata,'$.date') AS date, role, headline, substr(content, 1, 200) AS preview "
    "FROM hist.conversation_history WHERE kind='conversation' "
    "AND json_extract(metadata,'$.date') BETWEEN '2023-04-08' AND '2023-04-25' "
    "ORDER BY msg_index")
print(len(rows))
for r in rows:
    print(r['seq'], r['step_index'], r['date'], r['role'], (r['preview'] or '')[:120].replace('\n',' '))
\end{lstlisting}

\begin{lstlisting}[style=scroll, caption={Per-session aggregation of page 1 in the kernel.}, label={lst:sum-c1}]
rows_all = list(rows)
print(len(rows_all), rows_all[0]['seq'], rows_all[-1]['seq'], rows_all[-1]['date'])
# Aggregate per session: date range
from collections import defaultdict
sess = defaultdict(lambda: {'dates': set(), 'roles': 0})
for r in rows_all:
    sess[r['step_index']]['dates'].add(r['date'])
    sess[r['step_index']]['roles'] += 1
for s in sorted(sess):
    print(s, sorted(sess[s]['dates']), sess[s]['roles'])
\end{lstlisting}

\begin{lstlisting}[style=scroll, caption={Paging query; steps 3--5 rerun it changing only the \texttt{OFFSET}.}, label={lst:sum-c2}]
rows2 = ms.sql_query(
    "SELECT seq, step_index, json_extract(metadata,'$.date') AS date, role, substr(content,1,180) AS preview "
    "FROM hist.conversation_history WHERE kind='conversation' "
    "AND json_extract(metadata,'$.date') BETWEEN '2023-04-13' AND '2023-04-25' "
    "ORDER BY msg_index LIMIT 1000 OFFSET 0")
print(len(rows2))
rows2 = list(rows2)
from collections import defaultdict
sess2 = defaultdict(lambda: {'dates': set(), 'n': 0})
for r in rows2:
    sess2[r['step_index']]['dates'].add(r['date'])
    sess2[r['step_index']]['n'] += 1
for s in sorted(sess2):
    print(s, sorted(sess2[s]['dates']), sess2[s]['n'])
print(rows2[-1]['seq'], rows2[-1]['date'])
\end{lstlisting}

\begin{lstlisting}[style=scroll, caption={Kernel-side merge of all five pages, de-duplicated by \texttt{seq}.}, label={lst:sum-c6}]
# Merge all pages by seq (dedupe)
all_turns = {}
for coll in [rows_all, rows2, rows3, rows4, rows5]:
    for r in coll:
        all_turns[r['seq']] = r
merged = [all_turns[k] for k in sorted(all_turns)]
print(len(merged), merged[0]['seq'], merged[0]['date'], merged[-1]['seq'], merged[-1]['date'])
from collections import defaultdict
sess_all = defaultdict(lambda: {'dates': set(), 'n': 0, 'first_seq': None, 'last_seq': None})
for r in merged:
    d = sess_all[r['step_index']]
    d['dates'].add(r['date'])
    d['n'] += 1
    if d['first_seq'] is None:
        d['first_seq'] = r['seq']
    d['last_seq'] = r['seq']
for s in sorted(sess_all):
    print(s, sorted(sess_all[s]['dates']), sess_all[s]['n'], sess_all[s]['first_seq'], sess_all[s]['last_seq'])
\end{lstlisting}

\begin{lstlisting}[style=scroll, caption={Positional sampling---first 3 and last 2 user turns per session.}, label={lst:sum-c7}]
from collections import defaultdict
sess_turns = defaultdict(list)
for r in merged:
    sess_turns[r['step_index']].append(r)

# Print first 3 and last 2 user turns per session to get arc
for s in sorted(sess_turns):
    users = [r for r in sess_turns[s] if r['role'] == 'user']
    print(f"=== Session {s} ({users[0]['date']}) -- {len(users)} user turns ===")
    for r in users[:3] + users[-2:]:
        txt = r['preview'].replace('[Session ', '[S').replace('] user: ', '] U: ')
        print(f"  {r['seq']}: {txt[:160]}")
    print()
\end{lstlisting}

\begin{lstlisting}[style=scroll, caption={Head-only deep read (first 6 user turns) of sessions 61, 63, 66, 70, 76, 77, 80.}, label={lst:sum-c8}]
# Pull full user-turn content for decision-rich sessions and print to read
import json
def full_content(seq):
    rr = ms.sql_query("SELECT content FROM hist.conversation_history WHERE seq=?", (seq,))
    return rr[0]['content'] if rr else None

# Sample: first ~6 user turns of sessions 61,63,66,70,76,77,80
for s in [61,63,66,70,76,77,80]:
    ut = [r for r in sess_turns[s] if r['role']=='user']
    print(f"\n########## SESSION {s} ({ut[0]['date']}) ##########")
    for r in ut[:6]:
        c = full_content(r['seq'])
        print(f"\n--- seq {r['seq']} ---")
        print(c[:600])
\end{lstlisting}

\begin{lstlisting}[style=scroll, caption={The trajectory's only thematic search---growth vocabulary, bounded to the date range.}, label={lst:sum-c9}]
hits = []
for q in ["personal growth", "lessons learned", "personal growth trip", "growth", "changed me", "learned about myself"]:
    hits += ms.search(q, seq_range=(13611, 18058), scope='task', k=6)
seen = {}
for h in hits:
    if h['seq'] not in seen:
        seen[h['seq']] = h
print(len(seen))
for seq, h in sorted(seen.items()):
    print(seq, h.get('snippet','')[:160].replace('\n',' '))
\end{lstlisting}

\begin{lstlisting}[style=scroll, caption={Verbatim expansion of eight growth-evidence addresses.}, label={lst:sum-c10}]
rows = ms.expand([15138, 15489, 16441, 16678, 16824, 17216, 17712, 17932])
for r in rows:
    print(f"=== seq {r['seq']} | role={r['role']} ===")
    print(r['content'][:900])
    print()
\end{lstlisting}

\begin{lstlisting}[style=scroll, caption={Targeted expansions plus two budget searches.}, label={lst:sum-c11}]
rows = ms.expand([16009, 16018])
for r in rows:
    print(f"=== seq {r['seq']} | role={r['role']} ===")
    print(r['content'][:800])
    print()
# also search budget / Pittsburgh delay / food overspend
hits2 = ms.search("Pittsburgh delay", seq_range=(13611,18058), scope='task', k=5)
for h in hits2:
    print("HIT", h['seq'], h.get('snippet','')[:200].replace('\n',' '))
hits3 = ms.search("food overspend", seq_range=(13611,18058), scope='task', k=5)
for h in hits3:
    print("HIT", h['seq'], h.get('snippet','')[:200].replace('\n',' '))
\end{lstlisting}

\begin{lstlisting}[style=scroll, caption={Final point reads (\texttt{seq} 16685, 17712).}, label={lst:sum-c12}]
# Quick final check: confirm the "2 stops max" and "future travel" decision details from seq 17712 and 16678 context
r1 = ms.sql_query("SELECT content FROM hist.conversation_history WHERE seq=16685")
print(r1[0]['content'][:500])
r2 = ms.sql_query("SELECT content FROM hist.conversation_history WHERE seq=17712")
print(r2[0]['content'][:300])
\end{lstlisting}

\end{document}